\documentclass{article}

\usepackage{microtype}
\usepackage{graphicx}
\usepackage{subcaption}
\usepackage{booktabs} 

\usepackage{hyperref}

\usepackage[preprint]{icml2019}

\icmltitlerunning{Augmenting Experience via Teacher's Advice}

\usepackage[utf8]{inputenc} 
\usepackage[T1]{fontenc}    
\usepackage{hyperref}       
\usepackage{url}            
\usepackage{booktabs}       
\usepackage{amsfonts}       
\usepackage{nicefrac}       
\usepackage{microtype}      
\usepackage[english]{babel}
\usepackage{graphicx}
\usepackage[colorinlistoftodos]{todonotes}
\usepackage{color}
\usepackage{algorithmicx}
\usepackage{algorithm}
\usepackage{algpseudocode}%
\usepackage{subcaption}
\usepackage{multirow}
\usepackage{wrapfig}

\usepackage{amsmath}
\usepackage{graphicx}

\newcommand{\cut}[1]{}

\newcommand{\be}{\begin{equation}}
\newcommand{\ee}{\end{equation}}
\def\bea#1\eea{\begin{align}#1\end{align}}
\def\bean#1\eean{\begin{align*}#1\end{align*}}

\newcommand{\A}{\mathcal{A}}

\renewcommand{\L}{\mathcal{L}}

\newcommand{\E}{\mathbb{E}}
\newcommand{\ourmethod}{ACTRCE}

\newif\ifshort
\shortfalse

\ifshort
\usepackage{titlesec}
\titlespacing\section{0pt}{0pt plus 4pt minus 2pt}{0pt plus 2pt minus 2pt}
\titlespacing\subsection{0pt}{0pt plus 4pt minus 2pt}{0pt plus 2pt minus 2pt}
\else

\fi

\begin{document}

\twocolumn[
\icmltitle{ACTRCE: Augmenting Experience via Teacher’s Advice \\For Multi-Goal Reinforcement Learning}



\icmlsetsymbol{equal}{*}

\begin{icmlauthorlist}
\icmlauthor{Harris Chan}{equal,to,vec}
\icmlauthor{Yuhuai Wu}{equal,to,vec}
\icmlauthor{Jamie Kiros}{goo}
\icmlauthor{Sanja Fidler}{to,vec}
\icmlauthor{Jimmy Ba}{to,vec}
\end{icmlauthorlist}

\icmlaffiliation{to}{Department of Computer Science, University of Toronto, Toronto, Canada}
\icmlaffiliation{goo}{Google Brain, Toronto, Canada}
\icmlaffiliation{vec}{Vector Institute, Toronto, Canada}

\icmlcorrespondingauthor{Harris Chan}{hchan@cs.toronto.edu}
\icmlcorrespondingauthor{Yuhuai Wu}{ywu@cs.toronto.edu}

\icmlkeywords{Grounded Language, Reinforcement Learning, Sparse Reward, Machine Learning, ViZDoom}

\vskip 0.3in
]

\printAffiliationsAndNotice{\icmlEqualContribution} 

\begin{abstract}
Sparse reward is one of the most challenging problems in reinforcement learning (RL). Hindsight Experience Replay (HER) attempts to address this issue by converting a failed experience to a successful one by relabeling the goals. Despite its effectiveness, HER has limited applicability because it lacks a compact and universal goal representation. We present Augmenting experienCe via TeacheR's adviCE (ACTRCE), an efficient reinforcement learning technique that extends the HER framework using natural language as the goal representation. We first analyze the differences among goal representation, and show that ACTRCE can efficiently solve difficult reinforcement learning problems in challenging 3D navigation tasks, whereas HER with non-language goal representation failed to learn. We also show that with language goal representations, the agent can generalize to unseen instructions, and even generalize to instructions with \textit{unseen lexicons}. We further demonstrate it is crucial to use hindsight advice to solve challenging tasks, and even small amount of advice is sufficient for the agent to achieve good performance.
\end{abstract}
\section{Introduction}
 
    Many impressive deep reinforcement learning (deep RL) applications rely on carefully-crafted reward functions to encourage the desired behavior.  
However, designing a good reward function is non-trivial~\citep{Ng1999}, and requires a significant engineering effort. 
For example, even for the seemingly simple task of stacking Lego blocks, ~\cite{Popov2017DataefficientDR} needed 5 complicated reward terms with different importance weights. 
Moreover, handcrafted reward shaping~\citep{BartoShapeReward} can lead to biased learning, which may cause the agent to learn unexpected and undesired behaviors ~\citep{openaiBoat}. 
	
    One approach to avoid defining a complicated reward function is to use a \textit{sparse and binary} reward function, i.e., give only a positive or negative reward at the terminal state, depending on the success of the task. 
    However, the sparse reward makes learning difficult.

    Hindsight Experience Replay (HER) \citep{HER} attempts to address this issue. 
    The main idea of HER is to utilize failed experiences by substituting with an alternative goal in order to convert them to successful experiences. 
    For their algorithm to work, Andrychowicz et al. made the non-trivial assumption that for every state in the environment, there exists a goal which is achieved in that state. 
    As the authors pointed out, this assumption can be trivially satisfied by choosing the goal space to be the state space. 
    However, representing the goal using the enormous state space is very inefficient and contains much redundant information.
    
    For example, if we want to ask an agent to avoid collisions while driving where the state is the raw pixel value from the camera, then there can be many states (i.e. frames) that achieve this goal. It is redundant to represent the goal using each state.

    Therefore, we need a goal representation that is (1) expressive and flexible enough to satisfy the assumption in HER, while also being (2) compact and informative where similar goals are represented using similar features. 
    Natural language representation of goals satisfies both requirements. 
    First, language can flexibly describe goals across tasks and environments. 
    Second, language representation is abstract, hence able to compress any redundant information in the states. %
    Recall the previous driving example, for which we can simply describe ``avoid collisions" to represent all states that satisfy this goal. 
    Moreover, the compositional nature of language provides transferable features for generalizing across goals. 
    
   In this paper, we combine the HER framework with natural language goal representation, and propose an efficient technique called Augmenting experienCe via TeacheR's adviCE (ACTRCE; pronounced ``actress'') to a broad range of reinforcement learning problems. Our method works as follows. Whenever an agent finishes an episode, a teacher gives advice in natural language to the agent based on the episode. The agent takes the advice to form a new experience with a corresponding reward, alleviating the sparse reward problem. For example, a teacher can describe what the agent has achieved in the episode, and the agent can replace the original goal with the advice and a reward of 1. We show many benefits brought by language goal representation when combining with hindsight advice. The agent can efficiently solve reinforcement learning problems in challenging 2D and 3D  environments; it can generalize to unseen instructions, and even generalize to instruction with \textit{unseen lexicons}. We further demonstrate it is crucial to use hindsight advice to solve challenging tasks, but we also found that little amount of hindsight advice is sufficient for the agent's performance to significantly improve, showing the practical aspect of the method.

    We note that our work is also interesting from a language learning perspective. Learning to achieve goals described in natural language is part of a class of problem called language grounding \citep{HARNAD1990}, which has recently received increasing interest, as grounding is believed to be necessary for more general understanding of natural language. 
    Early attempts to ground language in a simulated physical world \citep{WINOGRAD1972,Siskind1994} consisted of hard coded rules which could not scale beyond their original domain. Recent work has been using reinforcement learning techniques to address this problem \citep{Deepmind3DLang,chaplot2017gated}. Our work combines reinforcement learning and rich language advice, providing an efficient technique for language grounding. 

\section{Background} 
	\subsection{Reinforcement Learning and Deep Q-Networks}
	We first review the traditional reinforcement learning setting, where an agent interacts with an infinite-horizon, discounted Markov Decision Process $(\mathcal{S},\mathcal{A},\gamma,P,r)$. Here, $\mathcal{S}$ is the state space, $\mathcal{A}$ is the action space, $\gamma$ the discount factor, $P$ the transition model and $r$ is a reward function $r: \mathcal{S}\times \mathcal{A} \to \mathbb{R}$. We consider a policy $\pi_\theta(a|s)$ parameterized by $\theta$. At time $t$, the agent chooses an action $a_t\in \mathcal{A}$ according to its policy $\pi_{\theta}(a|s_t)$ where $s_t\in \mathcal{S}$ is the current state. The environment in turn produces a reward $r(s_t,a_t)$ and transitions to the next state $s_{t+1}$ according to the transition probability $P(s_{t+1}|s_t,a_t)$. The goal of the agent is to maximize the expected $\gamma$-discounted cumulative return $\mathbb{E}_{\pi}[\sum_{t=0}^{\infty}\gamma^{t} r(s_{t},a_{t})]$ with respect to the policy parameters $\theta$. The action-value function is denoted as $Q^\pi(s_t,a_t)=\mathbb{E}_{\pi}[{\sum_{i=t}^\infty \gamma^{i-t} r(s_i, a_i)|s=s_t, a=a_t}]$, which is the expected discounted sum of rewards obtained by performing an action $a$ at state $s$ and following the policy $\pi$ thereafter.
	We denote $\pi^*$ as the \textit{optimal policy} such that $Q^{\pi^*}(s,a) \geq Q^{\pi}(s,a)$ for every $s \in \mathcal{S}$, $a \in \mathcal{A}$ and policy $\pi$. The \textit{optimal Q-function} $Q^*=Q^{\pi^*}$ satisfies the \textit{Bellman equation}:
    \begin{align}
    Q^{*}(s,a) = \E _{s' \sim p(\cdot | s,a)} \bigg[ r(s,a) + \gamma \max_{a' \in \A} Q^{*} (s', a')\bigg]
    \end{align} 
    Q-learning \citep{Sutton2018} is an off-policy, model-free RL algorithm that is based on the Bellman equation. The algorithm uses semi-gradient descent \citep{Sutton2018} to minimize the squared Temporal Difference (TD) error: $\L = \E(Q(s_t, a_t) - y_t)^2$, where the $y_t = r_t + \gamma \max_{a' \in \A} Q(s_{t+1}, a')$ is the \textit{TD target}.
    Deep Q-Network \citep{DQN} builds on the Q-learning algorithm and uses a neural network to approximate $Q^*$. It also uses a replay buffer as well as a target network to improve training stability.

    \subsection{Goal Oriented Reinforcement Learning}
    We review the goal-oriented reinforcement learning framework following \cite{UVFA}. We augment the previously defined infinite-horizon, discounted Markov Decision Process with a goal space $\mathcal{G}$. A goal $g$ is chosen from $\mathcal{G}$ and stays fixed for each episode. The goal induces a reward function $r_g: \mathcal{S}\times \mathcal{A} \to \mathbb{R}$, that assigns reward to a state conditioned on the given goal. At time step $t$, the agent $\pi(a_t|s_t,g)$ chooses an action conditioned on the current state $s_t$, as well as the current goal $g$. The agent's objective is to maximize the expected discounted cumulative return given the goal, i.e., $\mathbb{E}_{\pi}[\sum_{t=0}^{\infty}\gamma^{t} r(s_{t},a_{t},g)]$.

    \subsection{Hindsight Experience Replay (HER)}
      We follow \cite{HER} and consider a specific family of goal conditioned reward functions, where the reward is either $0$ or $1$, depending on the resulting state, $r_g(s_t, a_t) = f_g(s_{t+1})$, where $f_g:\mathcal{S}\to\{0,1\}$. The function $f_g$ acts as a predicate that determines whether the agent is successful or not according to $g$, and only assigns positive reward at the terminal state. Hence the reward function is very sparse, making it difficult for an agent to learn.
       
       HER proposes a solution to this problem. The agent first collects an episode of experiences $s_0,a_0,r_1,...,s_{T-1},a_{T-1},r_{T},s_T$ under the goal $g$, where $r_{T}=r_g(s_{T-1},a_{T-1})$ is the terminal reward. If $r_T$ is zero, one can replace $g$ with $g'\in\mathcal{G}$ such that $r_{g'}(s_{T-1},a_{T-1})=1$. With an off-policy algorithm, one is able to learn from the goal transformed experience.
      In order to flexibly relabel with a desirable goal, the authors assume that there exists a representation map from the state space to the goal space $\mathbb{T}: \mathcal{S} \to \mathcal{G}$, so that for each $s\in\mathcal{S}$, the corresponding goal representation satisfies the predicate $f_g\left(\mathbb{T}(s)\right)=1$. However, such mapping is not simple to construct. Although the author pointed out a trivial construction can be done by taking $\mathcal{G}=\mathcal{S}$ and $f_g(s) = [g=s]$, such goal representation is redundant and uninformative, and largely limits the applicability of the algorithm. Consider an example where the state is the first-person raw pixel observation in a 3-D environment, and the goal is to walk towards a particular object. There are many possible directions to approach the object from, which results in many different possible states that satisfy the goal. However, no subset of the raw pixel observation can abstractly represent the goal.
    
\section{Augmenting Experience via Teacher’s Advice (\ourmethod{})}

\begin{figure}[t]
        \centering
        \includegraphics[width=0.50\textwidth]{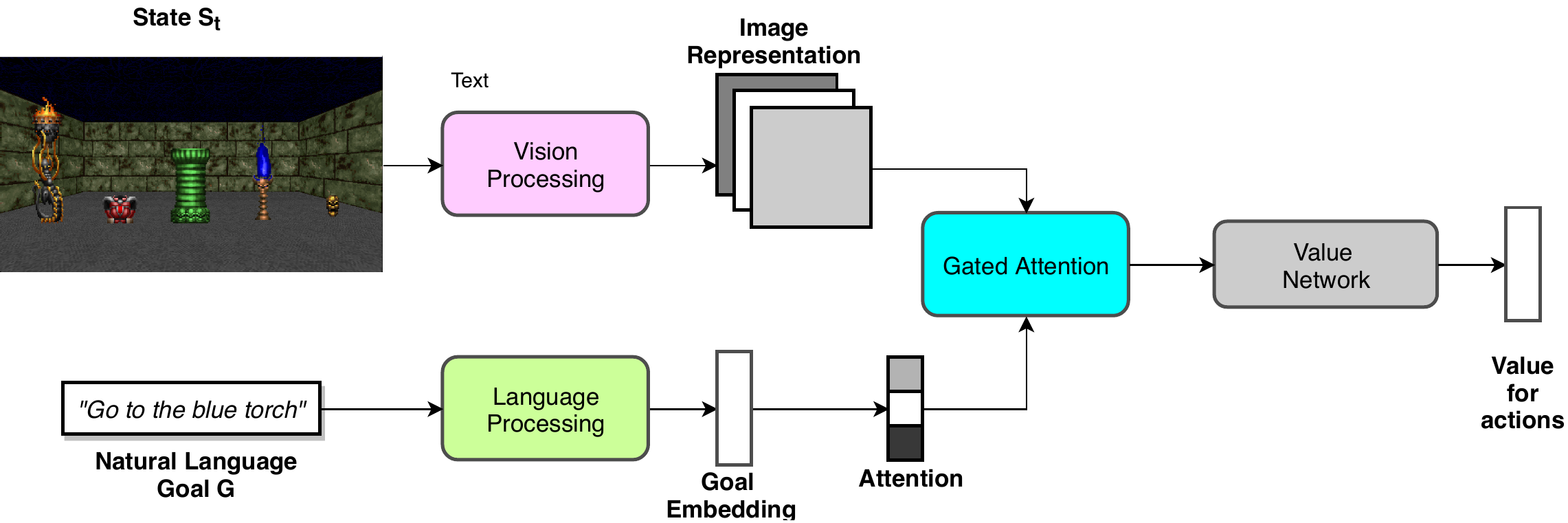}
        \caption{The diagram illustrates our model architecture.}
        \label{fig:arch_highlvl}
    \end{figure}
    For a goal oriented MDP, $\{\mathcal{S},\mathcal{G}, \mathcal{A},\gamma,P,r_g\}$, and a parameterized agent $\pi(a_t|s_t,g)$, our objective is to maximize the expected discounted cumulative return, i.e., $\mathbb{E}_{\pi}[\sum_{t=0}^{\infty}\gamma^{t} r_g(s_{t},a_{t})]$ for each $g \in \mathcal{G}$. Following HER, we assume there exists one or more goal representation mappings $\mathbb{T}: \mathcal{S} \to \mathcal{G}$, which describe what goal is satisfied at each state. For example, in the original HER paper, the mapping $\mathbb{T}$ is simply a subspace projection from the state. Such representation can work well for their robotic tasks, where subset of the state (i.e., the coordinates of objects and body parts) can reasonably well represent the goal. However, in general, such mapping can not abstract meaningful information about the goal, and will cause redundancy.
    
    Language provides an abstract representation for a goal and hence reduces redundancy in representation. This motivates our proposal to use \textit{natural language} for representing the goal space. Concretely, for each state $s\in\mathcal{S}$, we define $\mathbb{T}$ as a teacher that gives an advice $\mathbb{T}(s)$, a natural language description of the goal that is achieved in the state $s$. 
    To implement such a teacher, one can ask a human to look at a state and describe the goal in words, or generate a goal description automatically, as part of a narration or accessibility feature in a game. We take the latter approach, using the underlying emulator states to generate the teacher instruction. Given $\mathbb{T}$, we can convert a failure trajectory to a successful one by relabeling the original goal with the teacher advice.
    To illustrate, say the original goal for the episode is "Go to the armor", but at this particular time step, the agent has reached the blue torch instead. 
    The teacher then tells the agent that the goal that reflects the current state is "Go to blue torch". The agent can hence take the advice and use it as a positive experience.
    
    Besides positive reward signals, we also find that negative experience is necessary for training. Hence, in addition to a teacher who describes the achieved goal with a positive reward, we also introduce a teacher who gives negative reward with a description of a goal that has not been achieved.
    We further consider the scenario where there is more than one goal that can be satisfied in a state, e.g., the state that is described as ``Reach a blue torch" may also be described as ``Reach the largest blue object". Each different description of the goal corresponds to a different teacher (see more discussions and experiments on variations of teachers in Appendix \ref{sec:teacher_types}). Therefore, in general, there is a group of teachers of different kinds, each giving different advice. We then relabel the original goal with each advice and corresponding positive or negative reward, and augment the replay buffer with these trajectories. Algorithm~\ref{alg:herlang} describes our proposed approach formally. 
    
    Note that in the above formulation, we assume a MDP setting, and let the teacher give advice based solely on the terminal state $g'=\mathbb{T}(s_T)$. By contrast, in the POMDP setting, we ask the teacher to give advice based on the history of states and actions during the episode, i.e., $g'=\mathbb{T}(\{s_0, a_0,\dots, s_T\})$.
     
	\begin{algorithm}[H]
		\caption{Augmenting Experience via Teacher’s Advice (\ourmethod{}) }
		\label{alg:herlang}
		\begin{small}
		\begin{algorithmic}
			\State \textbf{Given}
			\begin{itemize}
				\item an off-policy RL algorithm $\mathbb{A}$, and replay buffer R. \Comment{e.g. DQN, DDPG, NAF, SDQN}
				\item A language goal space $\mathcal{G}$ and a desired goal space $\mathcal{G}_d$.
				\item A group of teachers $\{\mathbb{T}\}$
			\end{itemize}
			\For{episode\,=\,$1$,\,$M$}
			\State Sample desirable goal description $g\in\mathcal{G}_d$, initial state $s_0$.
			\For{$t\,=\,0,\,T-1$}
			\State Sample an action $a_t$ using behavioral policy from $\mathbb{A}$:
			\State \hspace{1cm} $a_t \leftarrow \pi_b(s_t || g)$ \Comment{$||$ denotes concatenation}
			\EndFor
			\For{every teacher $\mathbb{T}$}
			\State Compute advice $g'=\mathbb{T}(s_T)$. \Comment{Teacher's advice in natural language}
			\For{$t\,=\,0,\,T-1$}
			\State$r=r_{g'}(s_t, a_t)$
			\State Store transitions $\{(s_t||g',\,a_t,\,r,\,s_{t+1}||g')\}$ in $R$ 
			\EndFor
			\EndFor
			\State Perform training with $\mathbb{A}$ and a minibatch $B$ from the replay buffer.
			\EndFor
		\end{algorithmic}
		\end{small}
	\end{algorithm}

	\subsection{Language representation}
	There remains a question of how we deal with the natural language goal representation, a sequence of discrete symbols. In this paper, we explore two standard ways to convert a language goal into a continuous vector representation for neural network training. 
	One way is to represent each word as a one-hot vector, and use a recurrent neural network (RNN)
	to sequentially embed each word into its hidden state.
	We can then obtain some function of its hidden states (e.g., last hidden state, an attention over all hidden states) as the representation of the goal. The other way is to use a pre-trained language component obtained by other approaches and dataset (e.g. \cite{word2vec}) to represent the given language instructions. Since a pre-trained sentence embedding defines a reasonable similarity metric among sentences, one can expect the agent to understand \textit{unseen lexicons} that are closely related to the language goals in training. This allows the agent to gain better natural language understanding, and potentially be more robust to the language advice it gets from teachers (advice from humans can be quite noisy). 

	To integrate language representation of goals into the model, we design the architecture as follows. The architecture consists of 3 modules. One language component that takes an instruction and convert it into a continuous vector, and we apply an attention vector. The second component processes the observation to obtain a image representation using convolution neural networks. We then use gated attention from \cite{chaplot2017gated} to fuse goal information and the observation. The third component then takes the result of the fused representation and computes the output value. Figure.(\ref{fig:arch_highlvl}) shows the computational graph.


\section{Experiments}

In the following experiments, we demonstrate the effectiveness of the proposed method and discuss qualitative differences from the baseline. We first describe the experimental set up with our 2 environments, KrazyGrid World and ViZDoom. In the subsequent sections, we investigate:

\begin{enumerate}
     \item \textit{A comprehensive comparison between goal representations in hindsight advice, generalization, and semantic similarities.} We compare different goal representations: a naive one hot vector for each instruction, a Gated Recurrent Unit (GRU) \citep{GRU} that embeds the discrete language tokens, and a pre-trained word embedding. We show that as the number of instructions increases, the one hot approach does not scale as well as the GRU and pre-trained embedding. From visualizing the representation, we can see that structures emerge from GRU and pre-trained embeddings. We also demonstrate that pre-trained embedding representation can generalize to goals containing out of training vocabulary words. 
    \item \textit{Does the hindsight language advice help with learning? How much advice do we need?}  We show significant improvement in sample efficiency  by using advice from teachers. In many of the challenging tasks, the agent cannot learn at all without teachers' advice. We also show that significant improvement can be achieved even if we provide limited amount of teachers' advice, showing low burden of the method in practice.
\end{enumerate} 

	
\subsection{Environments and Training}
	We tested our method in a 2D grid world (KrazyGrid World \citep{EMAML}), as well as a 3D environment (ViZDoom \citep{chaplot2017gated}). 
	We describe the environment and our modifications below.
	
  	\vspace{-0.05cm}
	\textbf{KrazyGrid World (KGW)}~\citep{EMAML}: 
    KrazyGrid World is a 2D grid environment. For our experiments we chose to use 4 functionality of tiles: Goal, Lava, Normal, and Agent. We added an additional colour attribute in: Red, Blue, Green. The desired goals in this environment is to reach goals of different colours.
    Appendix \ref{app:KGW_goals} lists all language goals and additional environment details.
    
    \vspace{-0.1cm}
	
	\textbf{ViZDoom}\citep{Kempka2016ViZDoom,chaplot2017gated}: The 3D learning environment is based on the first person shooter game Doom. 
	The state is a raw pixel image from first person perspective of a room containing several random doom objects of various types, colours, and sizes. 
	 
	The goal for the episode is a natural language instruction in the form "Go to the [\textit{target attribute}] [\textit{object}]", such as "\textit{Go to the green torch}". 

	See Appendix \ref{app:vizdoom} for list of instructions and more description of ViZDoom.
    \vspace{-0.05cm}
    
    \textbf{Compositions of goals.} We refer language goals introduced in KGW and ViZDoom sections as singleton tasks. We further consider a set of more challenging tasks -- tasks that are composed of singleton tasks. Given two singleton tasks $A$, $B$, we take composition function ``and" and ``or" to form two new tasks ``$A$ and $B$", and ``$A$ or $B$". The task ``$A$ and $B$" is considered as complete when the agent completes both $A$ and $B$ within an episode, and ``$A$ or $B$" is considered as complete when one of the task is achieved. In our experiments, we consider all combinations of singleton tasks with both compositions ``and" and ``or" for KGW, and ``and" for ViZDoom.
    \vspace{-0.05cm}
    
    \textbf{Training details.}
    We used the DQN algorithm as the base reinforcement learning algorithm in both environments. A detailed architecture description for all of our experienments can be found in Appendix \ref{app:arch}. We carried out multi environment training as follows. First we sample 16 different environments. We collected data from all 16 environments. We updated the agents with an average gradient. We then resampled 16 environments. Further training details can be found in Appendix \ref{app:training}.

\subsection{Investigating different types of goal representation}
In this section, we investigate the hypothesis that using language for goal representation helps learning in various aspects. We compare language-based goal representations and a non-language goal representation. In summary, we show that when the difficulty of the tasks increases, language goal representations became more effective in providing learning signals. We also show how one can use language goal representation to generalize to unseen goals in training, whereas non-language goal representation could not. With a pre-trained sentence embedding, we can even generalize to instructions consisting of \textit{unseen lexicons}, showing the robustness of the language goal representation. Three goal representations are described as follows,

    \textbf{Language Sentence Representation with GRUs}: For each instruction, we represent each word as a one hot vector, and sequentially embed words into a GRU. We use take the last hidden state representation of the GRU for representing the instruction. We use the same GRU architecture from \cite{chaplot2017gated}, with hidden size 256.
    
    \textbf{Pre-trained Language Sentence Representation}: 
    We also consider models where the language component is pre-trained. 
    We use \textit{InferLite} \citep{kiros2018} as the pre-trained sentence representation and hold the parameters fixed during training. InferLite is a lightweight form of generic sentence encoder trained to perform natural language inference \citep{bowman2015large, N18-1101} resembling InferSent \citep{conneau-EtAl:2017:EMNLP2017} but without the use of RNNs. The original sentence embedding vector is of dimension 4096, and we use a learned linear layer to project it down to 256 for keeping every other part of the model same as the other two.
    
    \textbf{One-Hot Representation}: The non-language baseline represents each instruction as a one hot vector, which does not contain any language structure.
    We embed this one-hot goal representation to a vector with the same number of dimensions as the GRU representation. This embedding is learned independently for each goal encountered.

\subsubsection{How does each goal representation perform?}
\begin{figure}[t]
    \begin{center}
    \begin{tabular}{p{0.25\textwidth}p{0.25\textwidth}}
    \hspace{-0.5cm}\includegraphics[width=0.24\textwidth]{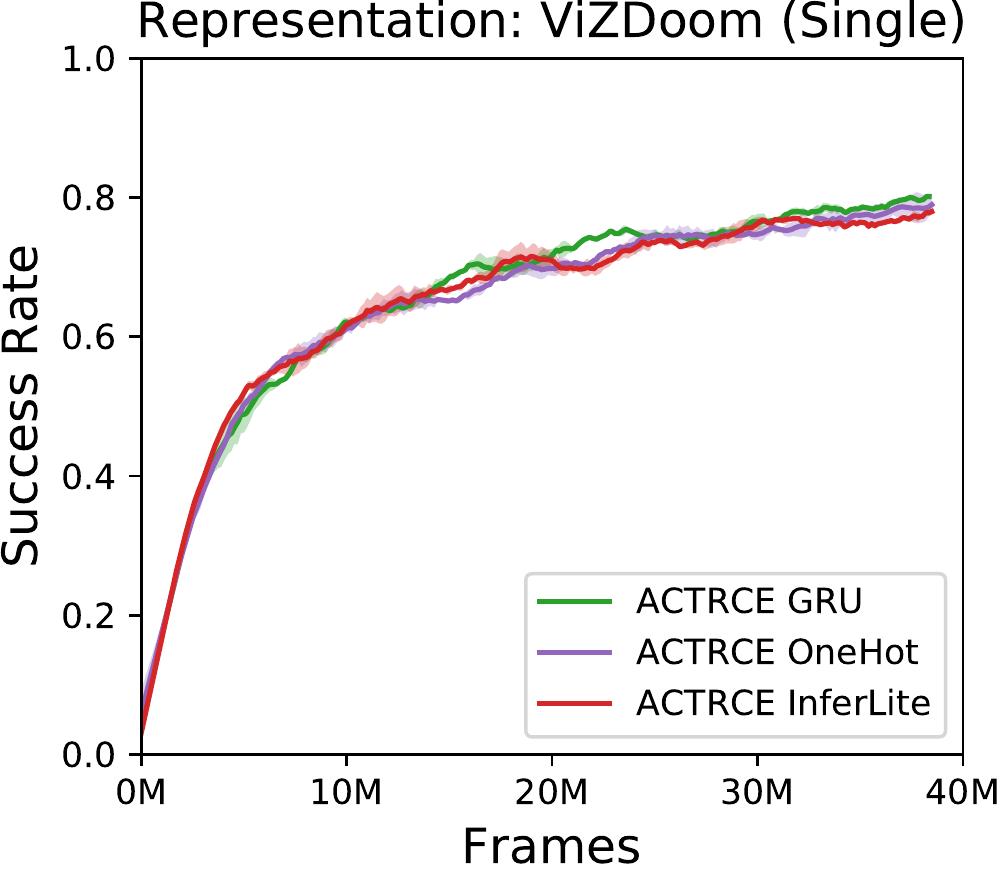}&
    \hspace{-0.5cm}\includegraphics[width=0.24\textwidth]{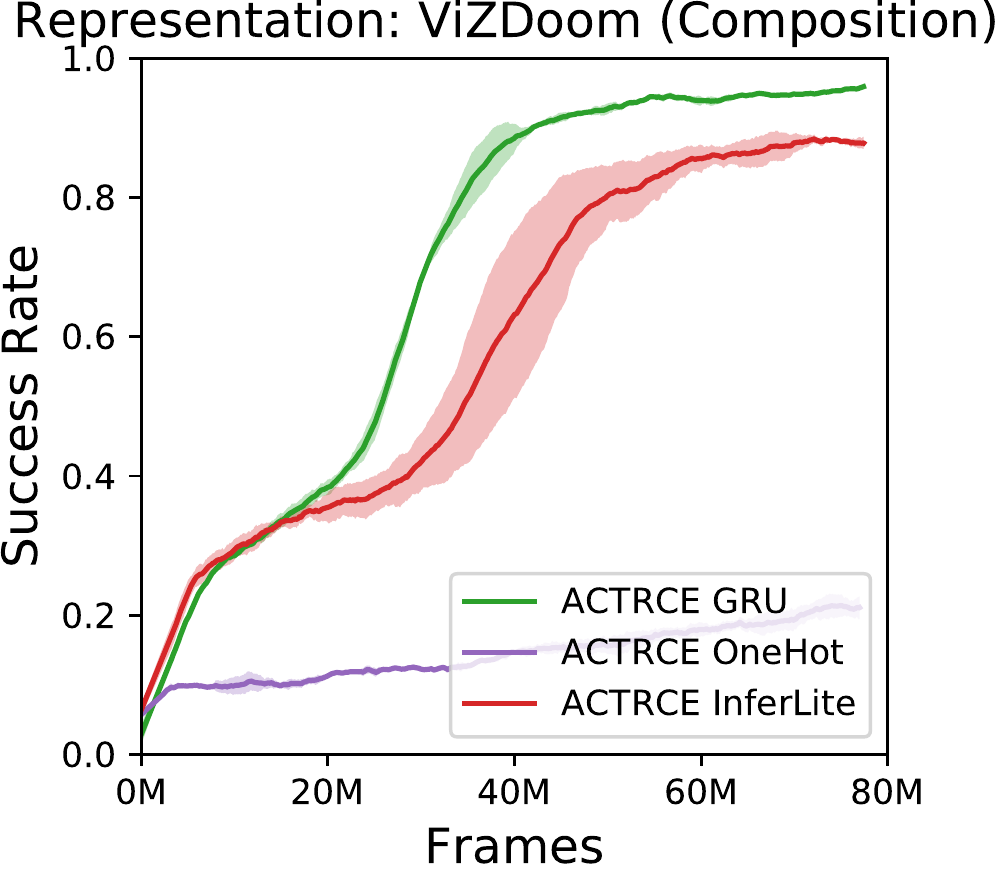} \\
    \centering(a)&\centering(b)
    \end{tabular}
    \caption{Performance in average success rate during training, comparing between different sentence embedding methods for the single target and composition task on ViZDoom.} 
    \label{fig:vizdoom_sentrep}
    \vspace{-0.3cm}
    \end{center}
\end{figure}

We use all three goal representations with hindsight advice for learning both singleton tasks and compositional tasks of ViZDoom environment. The singleton tasks consists of 7 objects, and the compositional tasks consists of 5 objects. The result of comparisons is plotted in Figure \ref{fig:vizdoom_sentrep}. We see that in an easier benchmark---singleton---tasks, agents using one-hot goal representations are still able to learn as quickly as agents using the other two goal representations. However, in a much more challenging benchmark--- compositional--tasks, the one-hot representations can only achieve a $24\%$ success rates, whereas ACTRCE with language representations can achieve $97\%$. Since one hot representations represent each goal independently, the agent is unable to generalize to any new unseen instructions. With language representation, we can easily generalize to \textit{unseen instructions} that consists of \textit{seen lexicons}. We reported the generalization results in Table \ref{tab:vizdoom_results}, under ``ZSL" (zero-shot learning). Remarkably, with GRU language goal representation, the agent is able to generalize to unseen instructions 83\% of the time, showing great generalization ability of language goal representations.

\subsubsection{Visualization of learned embeddings}
\begin{figure}[t!]
   \begin{center}
   \begin{tabular}{p{0.15\textwidth}p{0.15\textwidth}p{0.15\textwidth}}\\
   \hspace{-0.4cm}\includegraphics[width=0.15\textwidth]{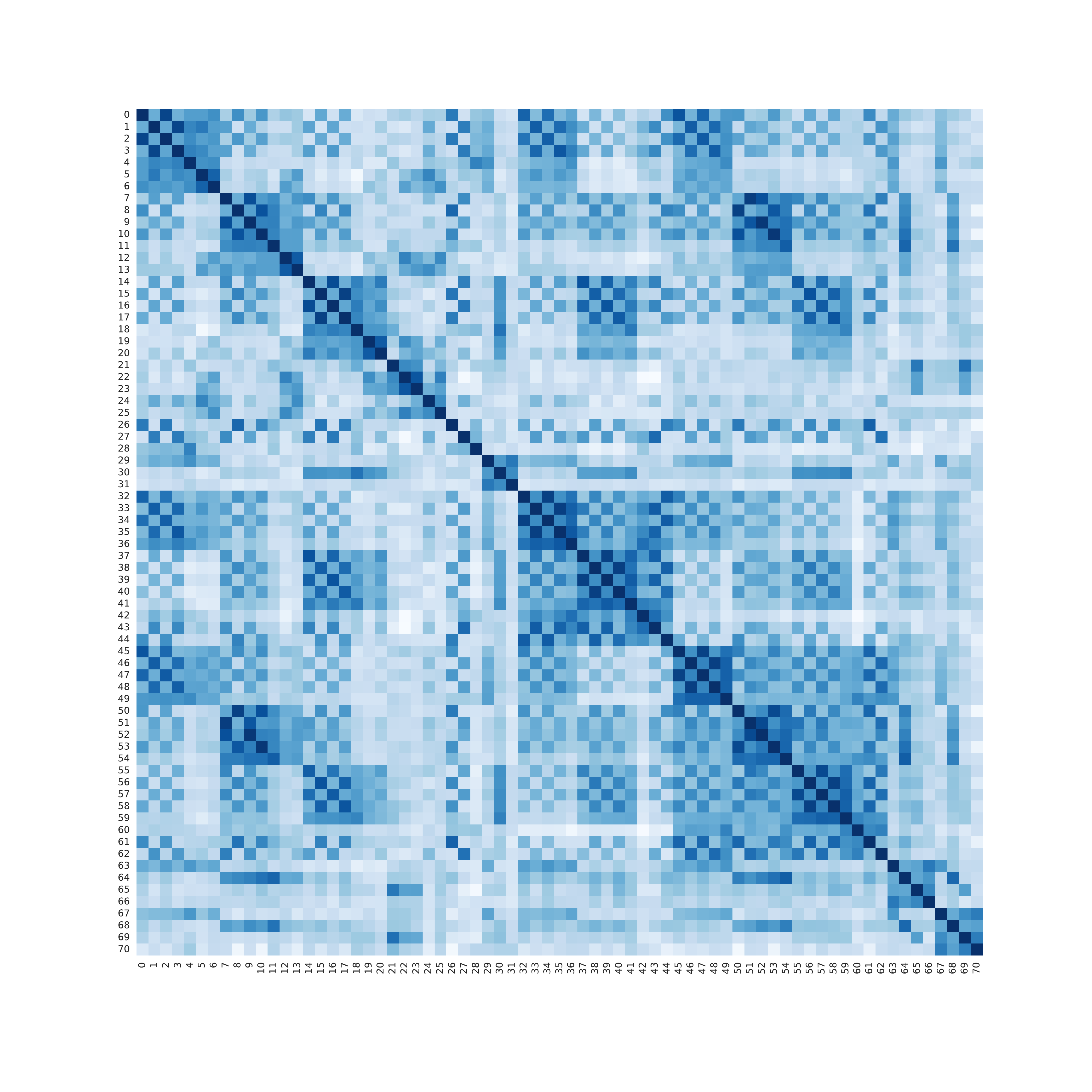}&
    \hspace{-0.4cm}\includegraphics[width=0.15\textwidth]{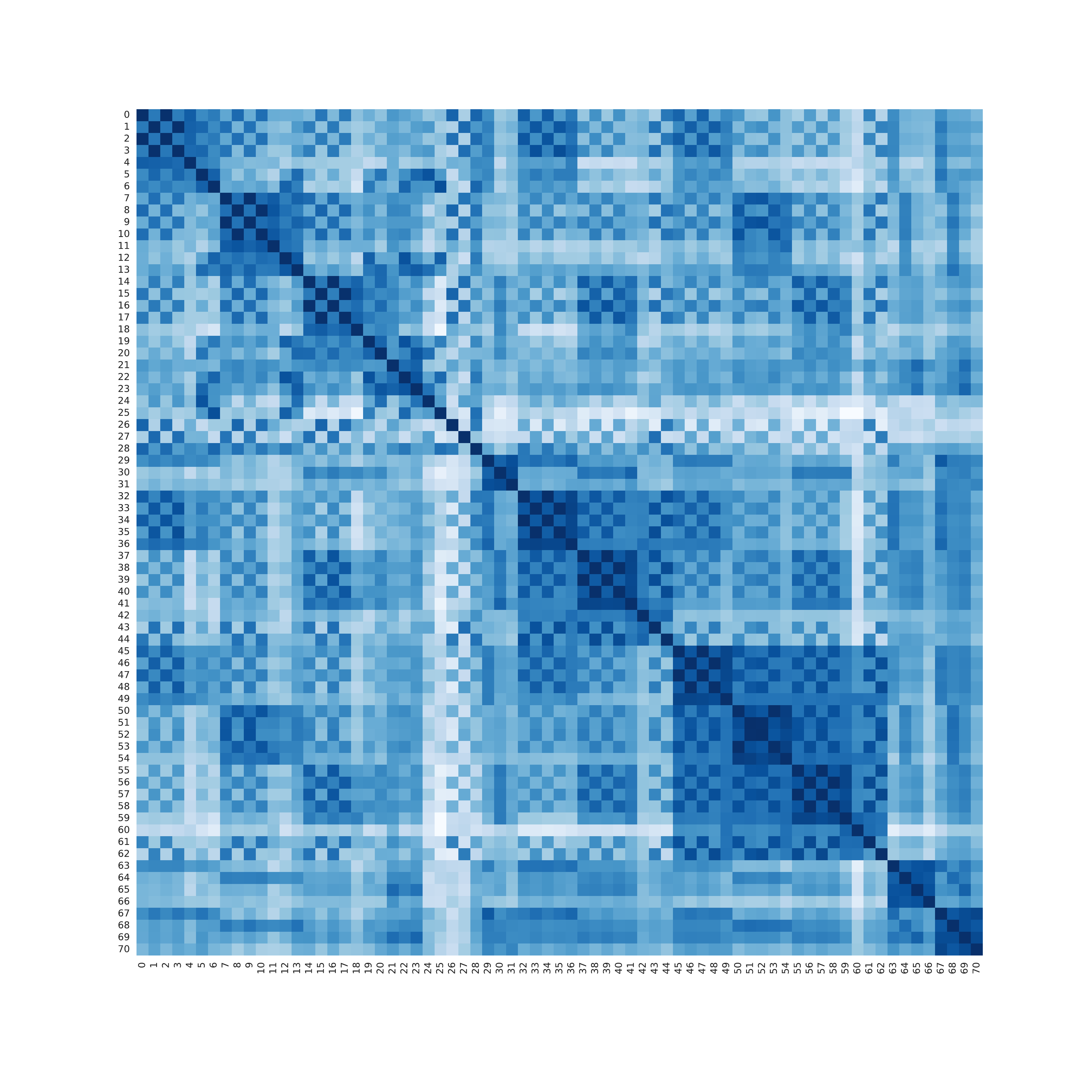}& 
    \hspace{-0.4cm}\includegraphics[width=0.15\textwidth]{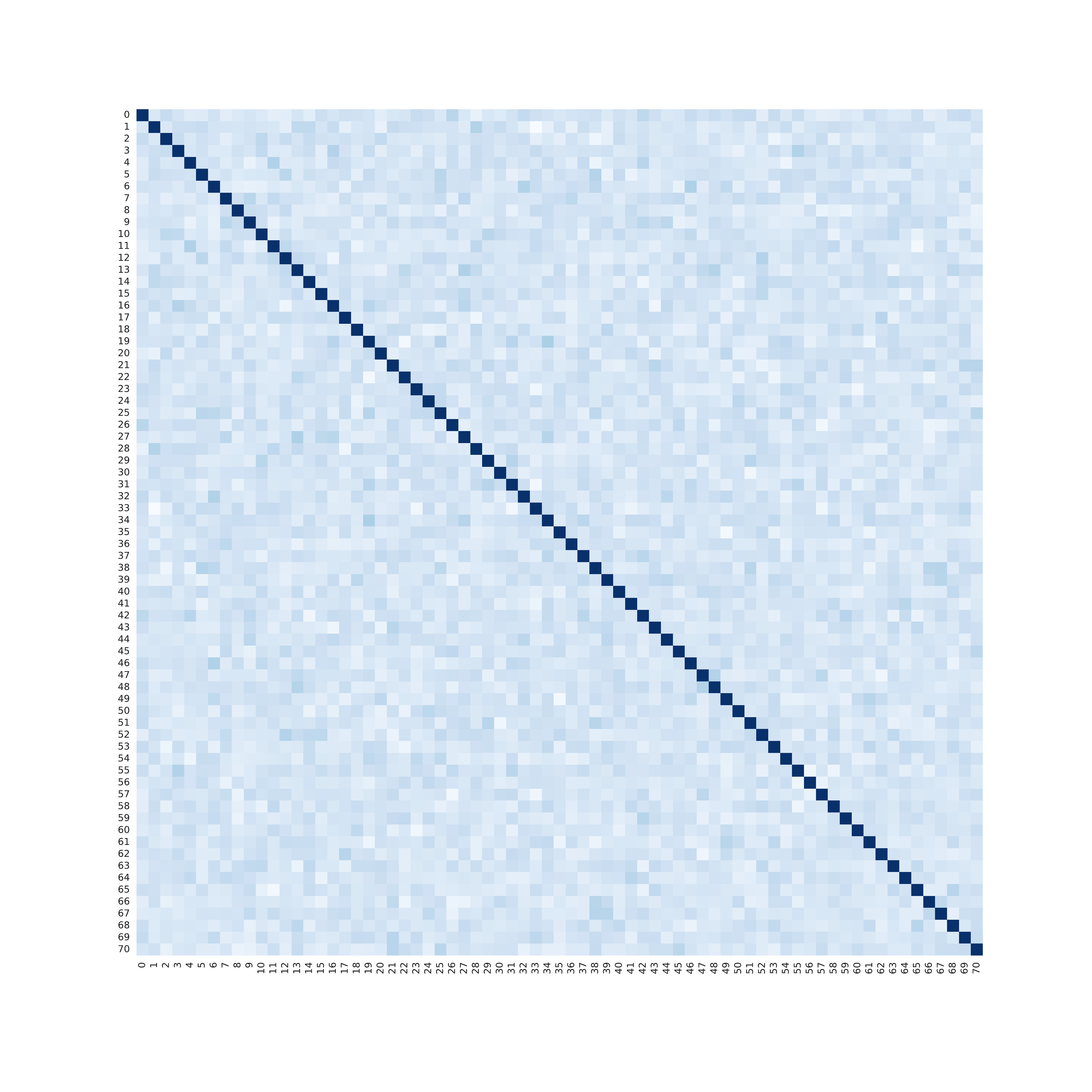}\\
    \hspace{-0.25cm}(a) \small GRU & \hspace{-0.25cm}(b)\small InferLite &\hspace{-0.25cm}(c)\small One-hot
    \end{tabular}
    \caption{Comparison among different sentence embedding methods showing the pairwise correlation between the sentence embedding vectors for each of the singleton instructions. The darker the colour, the higher the correlation.} %
    \label{fig:vizdoom_embedding_correlation}
    \vspace{-0.5cm}
    \end{center}
\end{figure}

We carried out a visualization analysis to see the statistical relations between the learned embeddings of goals. We took the model trained in the singleton VizDoom benchmark, where the agent was able to learn with all three goal representations. However, we find there are huge differences in learned embeddings. For each goal instruction, we extract the corresponding learned embedding for all three goal representation, which is the continuous vector before the attention layer (recall Fig.(\ref{fig:arch_highlvl})), all of size 256. We then calculated the correlation matrix for each goal representations. We plot all three correlation matrices in Figure.(\ref{fig:vizdoom_embedding_correlation}). The rows and columns are grouped by object type, then colour and size. We found that GRU and InferLite embeddings have very similar block-like structure, due to clustering of colours and shapes, while each one-hot goal embeddings share almost no correlation between each other. We further performed t-SNE \citep{maaten2008visualizing} embeddings and observe meaningful clustering with language goal representations and sporadic embeddings for one-hot goals. See more details in Appendix \ref{app:visualization_correlation}.

\setlength{\tabcolsep}{0.5em}
    \begin{table*}[t!]
    	\centering
    	\begin{tabular}{@{}ccccccc@{}}
    		\toprule
    		\multicolumn{2}{c}{\multirow{2}{*}{Method}}                                                 		 & \multicolumn{2}{c}\bf{Single (7 objects)}  & \multicolumn{2}{c}\bf{Composition (5 objects)}\\
    		\multicolumn{2}{c}{}                                                                               & MT          & ZSL   & MT          & ZSL      \\ \midrule
 		    \multirow{3}{*}
    		& DQN (GRU)  & ${0.08 \pm 0.01}$        & $ 0.065 \pm 0.007$          & $0.115 \pm 0.007$            & $0.07 \pm 0.04$  \\
    		& DQN (InferLite)  & ${0.05 \pm 0.07}$        & ${0.04 \pm 0.06}$          &       $ 0.10 \pm 0.02$      & $0.13 \pm 0.07 $  \\
    		& DQN (OneHot)  & ${ 0.035 \pm 0.007}$        & -          & $0.02 \pm 0.03$            & -  \\
    		\midrule
    		\multirow{3}{*}
    		& \ourmethod{}  (GRU)  & $\bf{0.80 \pm 0.06}$        & $\bf{0.76 \pm 0.03}$       & $\bf{0.97 \pm 0.04}$            & $\bf{0.83 \pm 0.09}$      \\ 
    		& \ourmethod{}  (InferLite)  & $\bf{0.80 \pm 0.01}$       &   $\bf{0.76 \pm 0.02}$    &  ${0.88 \pm 0.05}$            &  ${0.62 \pm 0.01}$     \\ 
    		& \ourmethod{}  (One Hot)  & $\bf{0.80 \pm 0.03}$        &    -    &   ${0.24 \pm 0.05}$          & -      \\ 
    		\bottomrule
    	\end{tabular}%
    	\caption{The table shows the averaged success rates, calculated over 100 episodes, in multitask and zero-shot scenario for the model trained with DQN versus DQN with \ourmethod{}. The standard deviations are calculated across 2 seeds. In the Multitask (\textbf{MT}) scenario, the goal for the episode is sampled from the training set of instructions but with a random environment initialization. For Zero-Shot (\textbf{ZSL}), the instruction is sampled from the held out test instruction set and with a random environment initialization.}
    	\label{tab:vizdoom_results}
    	\vspace{3pt}
    	
  		\vspace{-0.5cm}
    \end{table*}

\subsubsection{Generalizing to unseen lexicons with pre-trained word embedding}
We now show how to use pre-trained embeddings to allow our model to generalize to unseen lexicons at test time through representation transfer, similar in spirit to DeViSE \citep{frome2013devise} that was used for image classification with unseen classes. We took the agent trained in singleton tasks with InferLite goal representations, and replaced one word in each original instruction with its synonym\footnote{See Appendix \ref{sec:synonyms} for details on the synonyms used.}. We evaluate the performance of the agent on these new instructions that contain unseen lexicons. The results are shown in Tab.(\ref{tab:vizdoom_syn_list}). We find the agent is able to achieve tasks above $66\%$ of time. The ability of understanding sentences of similar meanings become useful when one implements the method with advice comes from humans. Since humans can describe the same meaning in many different ways, understanding synonyms can improve the robustness of learning in general noisy settings.
\setlength{\tabcolsep}{0.25em}
\begin{table}[h!]
\centering
\begin{tabular}{ccc}
\toprule
\textbf{Method}                   & \textbf{MT + Synonym} & \textbf{ZSL + Synonym} \\ \hline
\multirow{1}{*}
ACTRCE (InferLite)   & $\bf{0.66 \pm 0.05}$ & $\bf{0.62 \pm 0.02}$  \\
\bottomrule
\end{tabular}
\caption{Multitask evaluation with synonym and Zero-shot with synonym using InferLite.}
\label{tab:vizdoom_synonym}
\vspace{3pt}
\vspace{-0.5cm}
\end{table}

	\subsection{Do we need hindsight advice?}
	In the previous section, we demonstrate the effectiveness of language goal representation from various perspectives. In this section, we show how hindsight advice plays an important role in learning. We compared our method (denoted as ``\ourmethod{}") to the algorithm DQN, the base algorithm without any hindsight language advice. We show that without hindsight advice, DQN is not able to learn many challenging tasks. However, we also found that little advice (1\%) is sufficient for learning to take off, showing the practicality of our method. For language representation, we chose to use recurrent neural networks for embedding language goals, for both our method and the baseline method.
	
	\subsubsection{Comparison of ACTRCE to DQN}
\begin{figure*}[t!]
        \begin{center}
        \begin{tabular}{p{0.3\textwidth}p{0.3\textwidth}p{0.3\textwidth}}
        \hspace{-0.2cm}\includegraphics[width=0.3\textwidth]{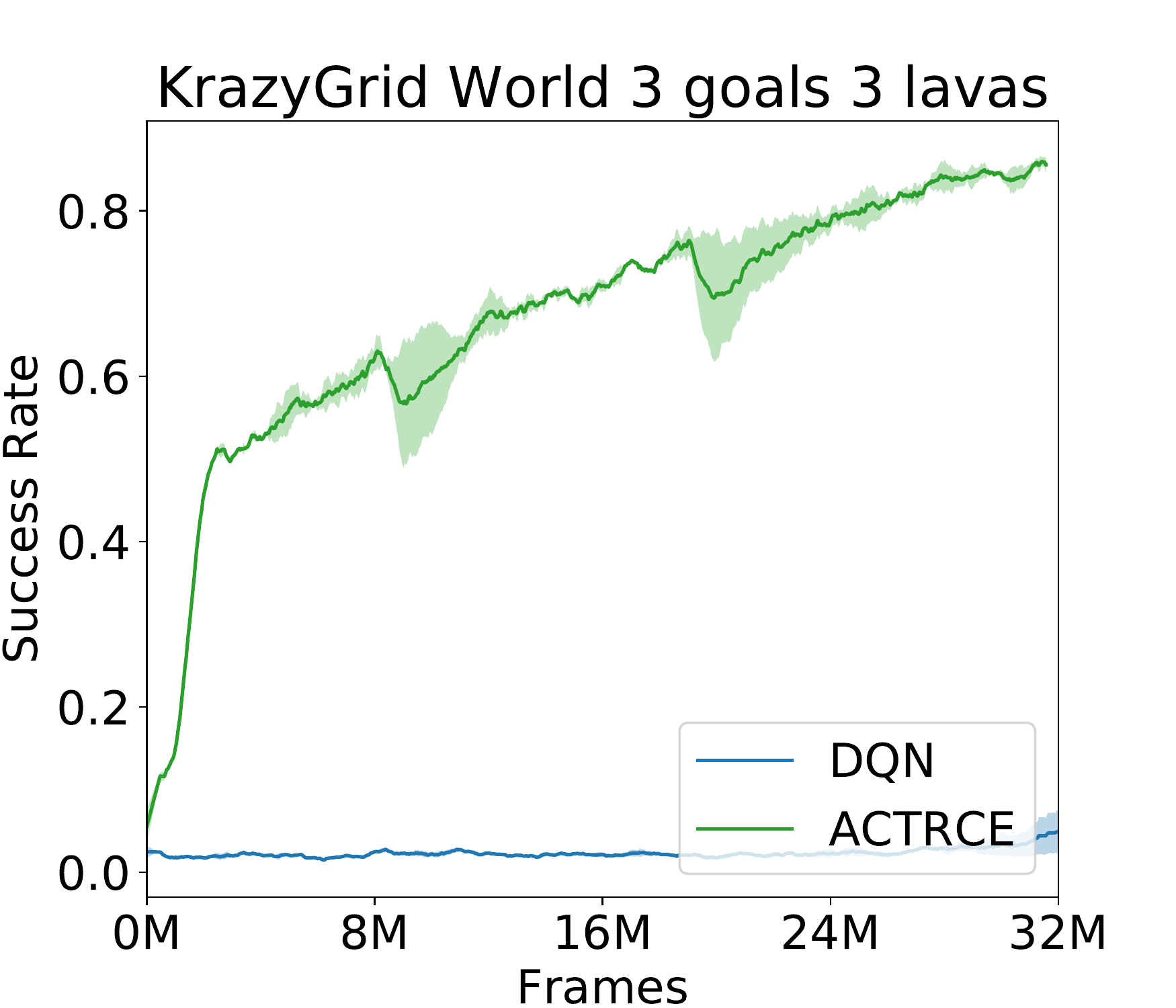}&
        \hspace{-0.2cm}\includegraphics[width=0.3\textwidth]{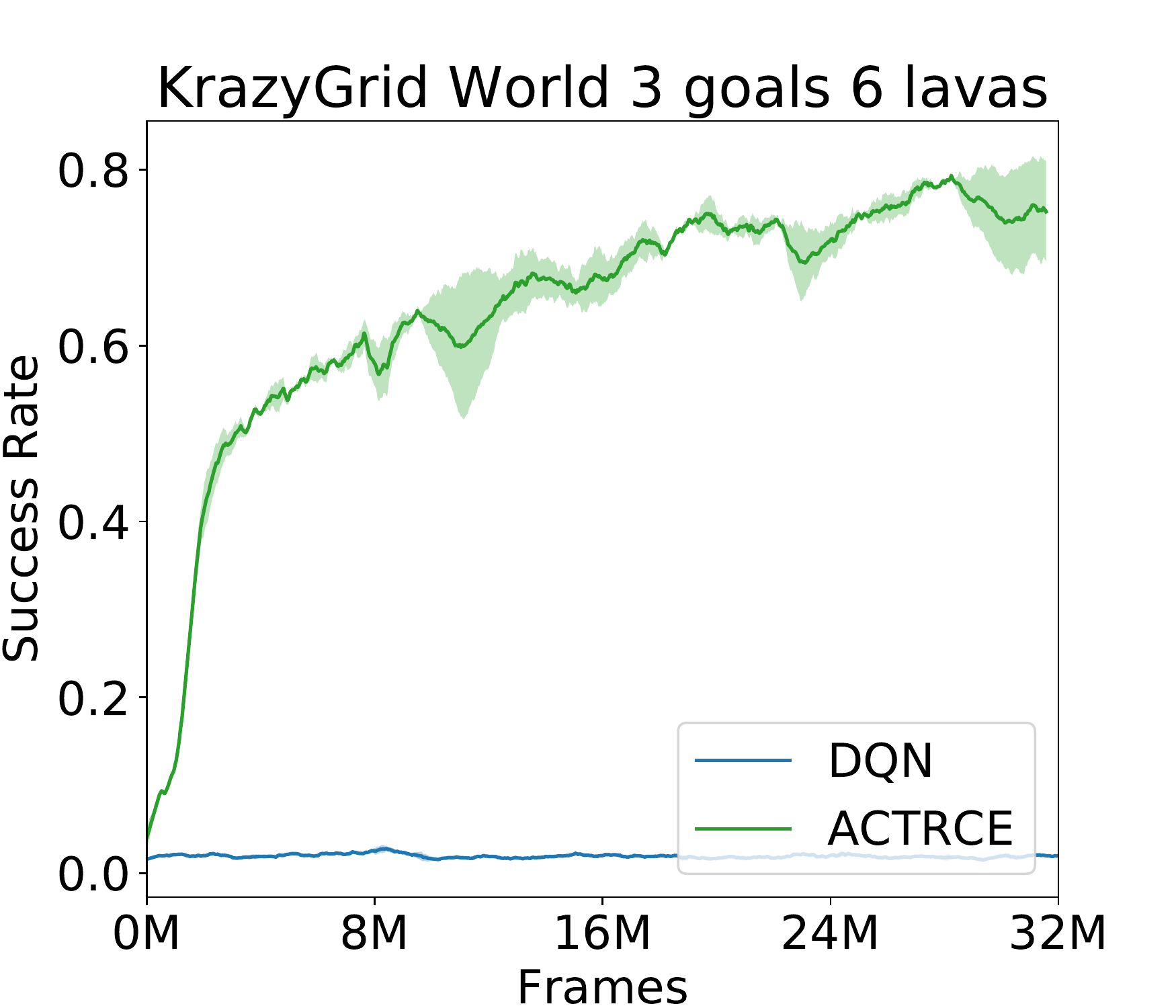}&
        \hspace{-0.2cm}\includegraphics[width=0.3\textwidth]{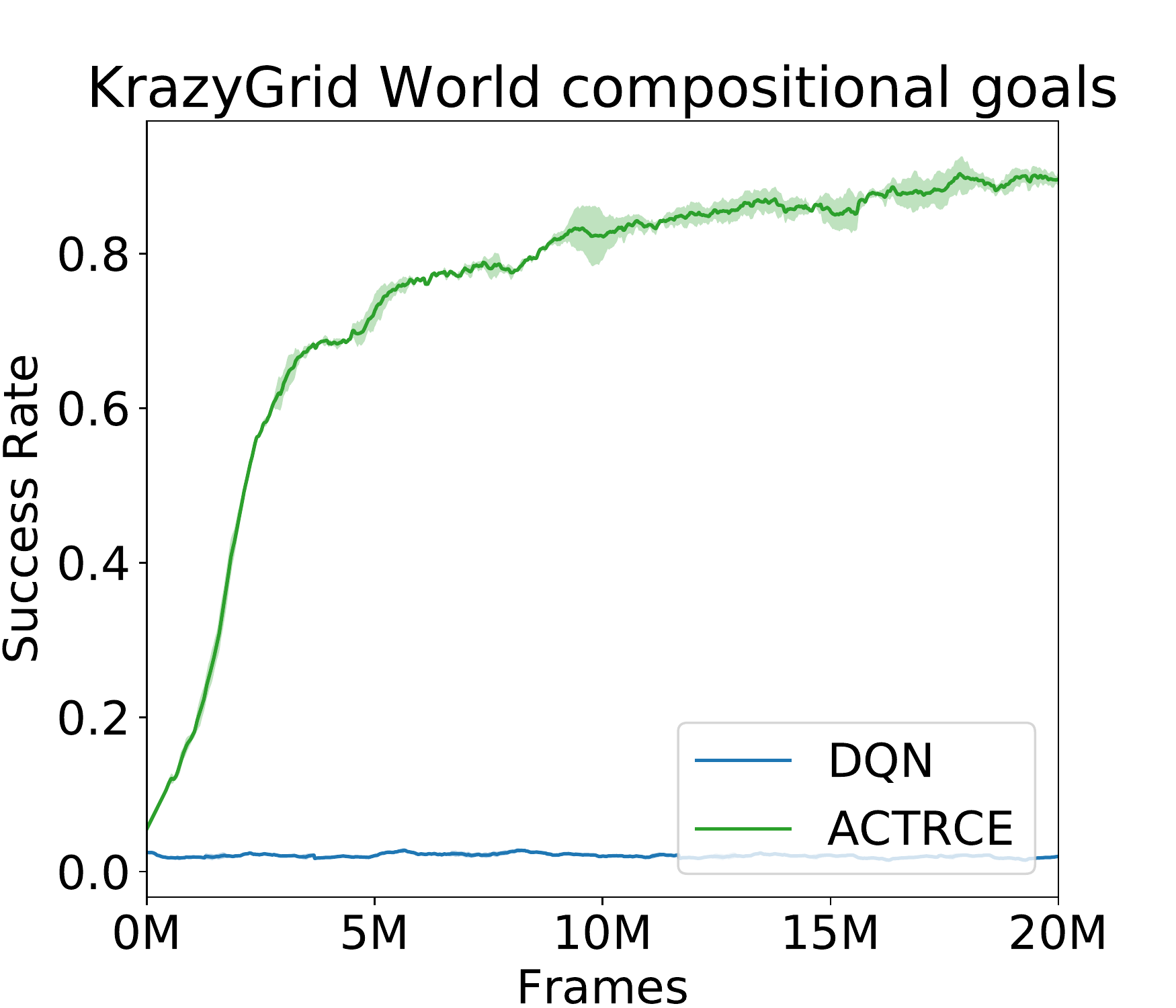}\\
        \hspace{2.0cm}(a)&\hspace{2.0cm}(b)&\hspace{2.0cm}(c)\\
        \hspace{-0.2cm}\includegraphics[width=0.3\textwidth]{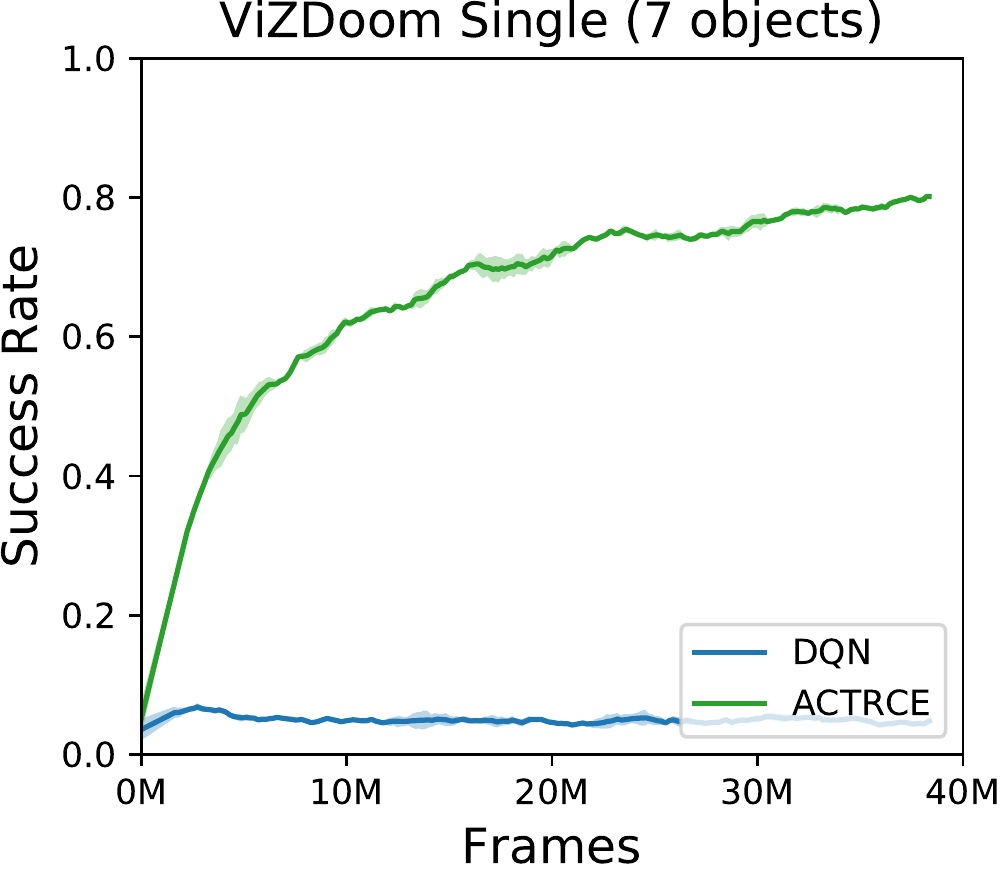}&
        \hspace{-0.2cm}\includegraphics[width=0.3\textwidth]{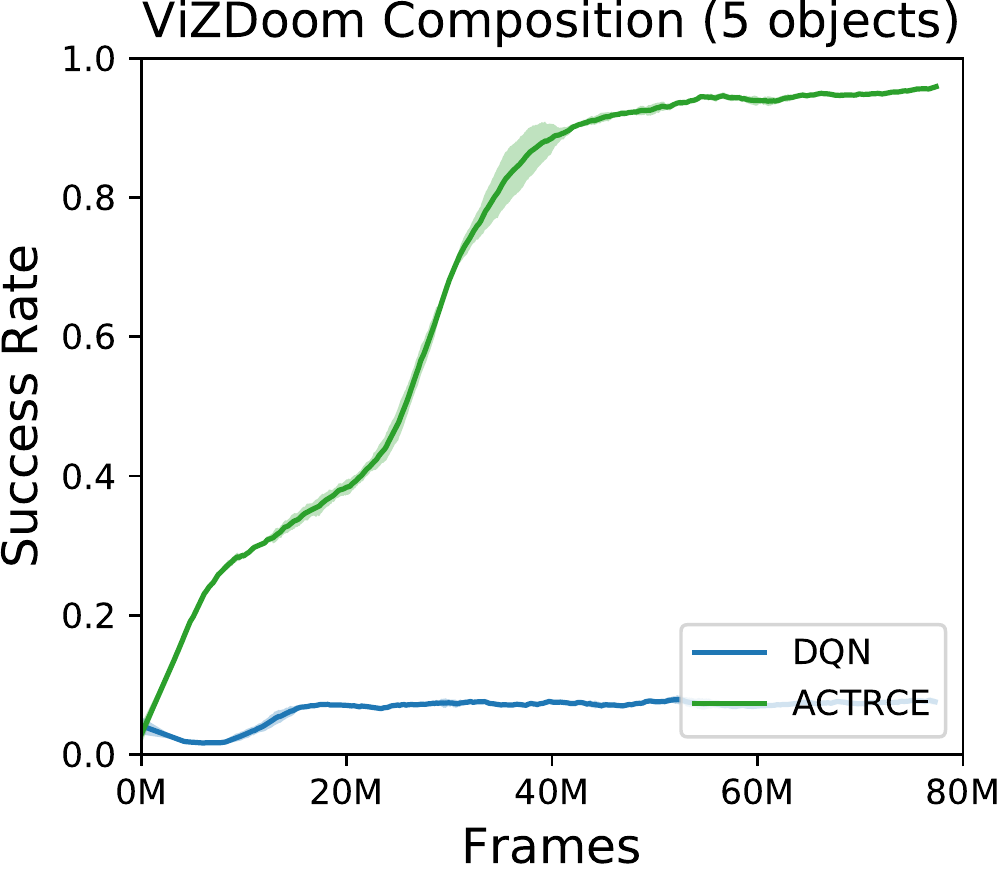}&
        \hspace{-0.2cm}\includegraphics[width=0.3\textwidth]{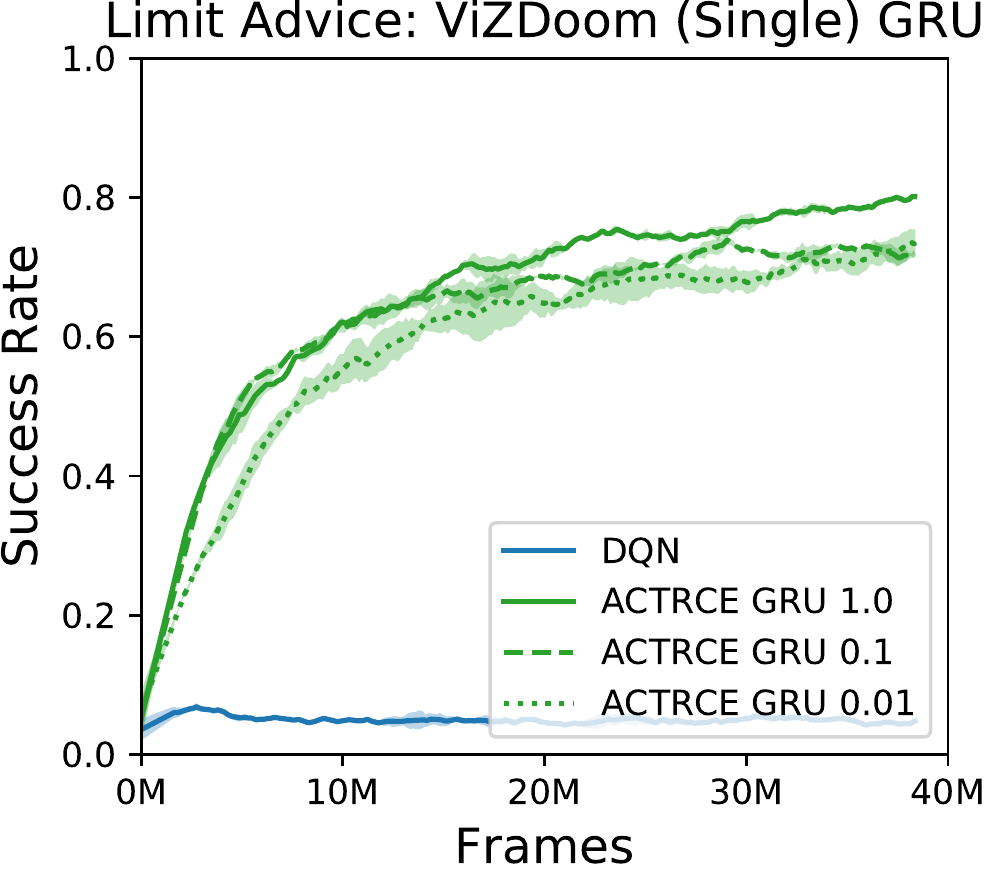}\\
        \centering(e)&\centering(f)&\centering(g)
        \end{tabular}
        \caption{Performance comparisons on KGW and ViZDoom environments. The success rates are calculated over all desired goals and 16 different environments. Shaded area represents standard deviation 2 random seeds.}
        \label{fig:Kgw_result}
        \vspace{-0.5cm}
        \end{center}
    \end{figure*}

   \paragraph{Singleton tasks.}
    For experiments on KGW, we tried two different kind of grids, one with 3 lavas and the other with 6 lavas, and both with 3 goals of different colours. Fig. \ref{fig:Kgw_result} (a) and (b) show the average success rate over all goals on 16-environments training. The shaded area represents the standard deviation over 2 random seeds. 
    The baseline DQN is shown in blue curve, which failed to learn at all. Our method (shown in green) quickly learned and achieved good results on both environments (around $80\%$ success rate). 
    
\begin{figure}[h!]
        \begin{center}
        \begin{tabular}{p{0.25\textwidth}p{0.25\textwidth}}
        \hspace{-0.2cm}\includegraphics[width=0.25\textwidth]{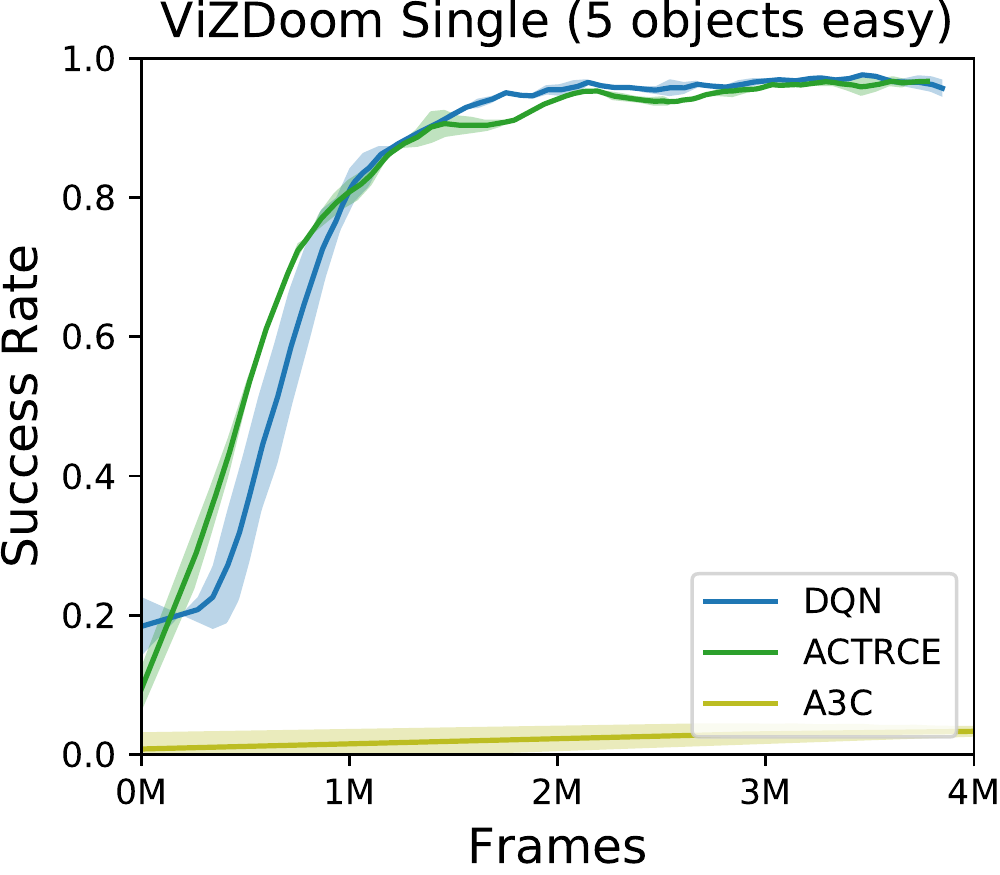} &
        \hspace{-0.2cm}\includegraphics[width=0.25\textwidth]{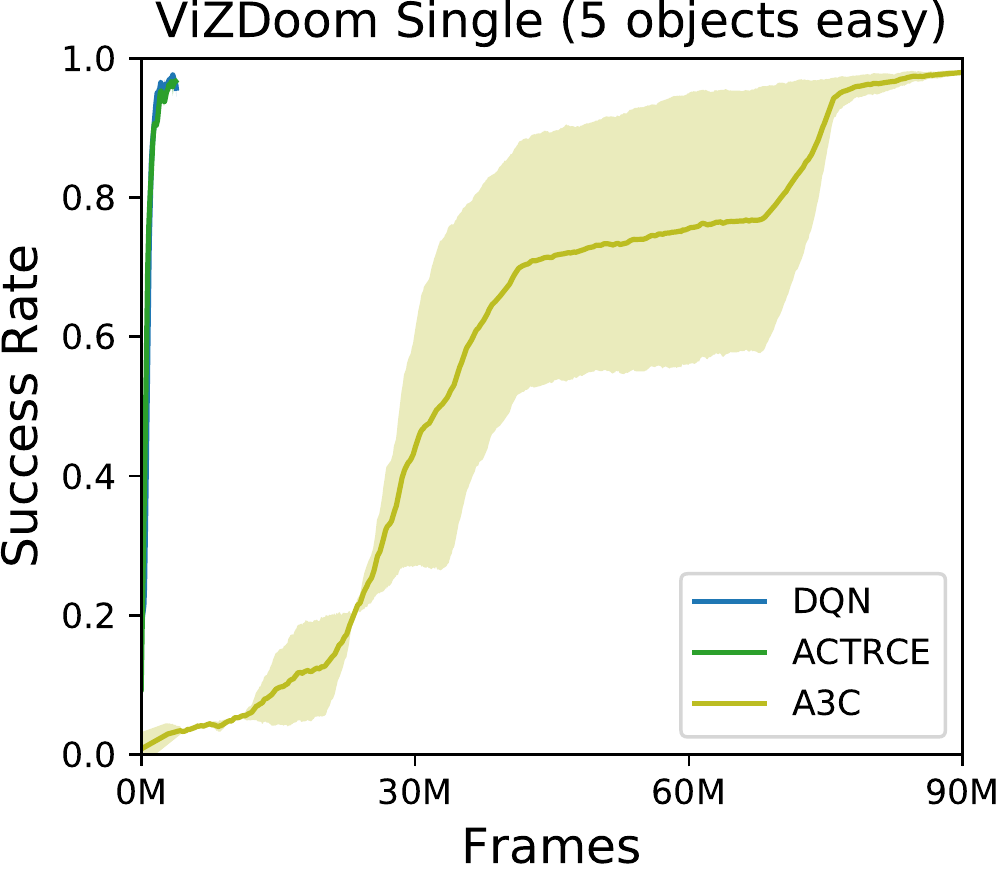}\\
        \hspace{-0.0cm}\centering(a)&\hspace{-0.0cm}\centering(b)
        \end{tabular}
        \caption{ViZDoom experiment with 5 objects in easy mode for single target case.}
        \label{fig:vizdoom_5obj_easy_single}
        \end{center}
    \end{figure}

\begin{figure}[h!]
        \begin{center}
        \begin{tabular}{p{0.25\textwidth}p{0.25\textwidth}}
        \hspace{-0.2cm}\includegraphics[width=0.25\textwidth]{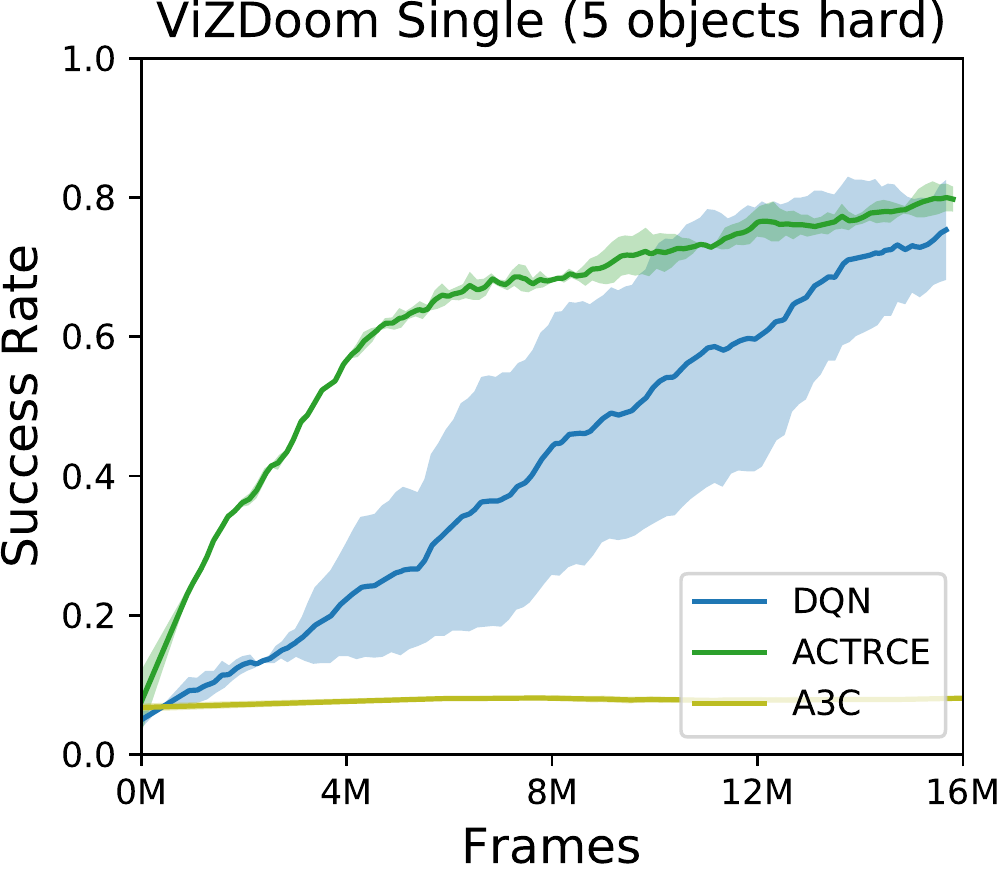} &
        \hspace{-0.2cm}\includegraphics[width=0.25\textwidth]{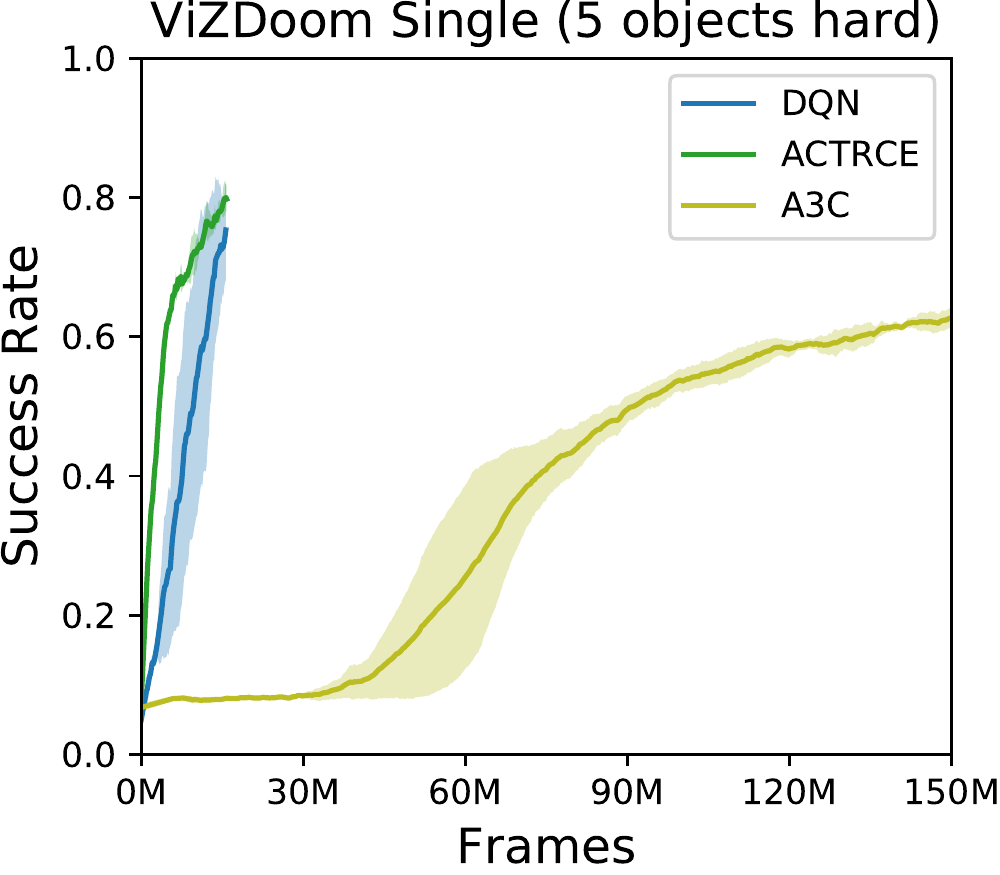}\\
        \hspace{-0.0cm}\centering(a)&\hspace{-0.0cm}\centering(b)
        \end{tabular}
        \caption{ViZDoom experiment with 5 objects in hard mode for single target case. }
        \label{fig:vizdoom_5obj_hard_single}
        \end{center}
    \end{figure}

    In ViZDoom experiments, we trained our agent in 3 configurations: 5 objects in (1) \textit{easy} and (2) \textit{hard} mode, and (3) 7-objects in the \textit{hard} mode with 50\% larger room size.  
    We ran the A3C baseline from \cite{chaplot2017gated}, from their implementation available online \cite{DeepRLGrounding} using the hyperparameter settings stated in their paper. Table \ref{tab:vizdoom_results_5obj_single} and Figure \ref{fig:vizdoom_5obj_easy_single} and \ref{fig:vizdoom_5obj_hard_single} summarize the training results for the 5 objects easy/hard mode. On the easy task, we were able to solve the task with A3C, but with an order of magnitude less sample efficiency compared to our DQN/\ourmethod{} implementation. On the hard task, we were unable to reproduce their results with our limited computational budget. A similar trend of an order of magnitude difference in sample efficiency is observed between A3C and DQN/ACTRCE on the hard task. 
      We found that in the more difficult 7-objects environment, only the agent trained with \ourmethod{} was able learn, compared to DQN. Figure \ref{fig:Kgw_result} (e) illustrates the  training instruction average success rate (over 100 episode chunks) vs. the number of frames set, when using \ourmethod{} versus DQN baseline. 
    Table \ref{tab:vizdoom_results} summarizes the agent's Multitask and Zero-Shot generalization performance. 
    
    \setlength{\tabcolsep}{0.4em}
    \begin{table*}[h!]
        \centering
        \begin{tabular}{l|cccc|ccc}
		\toprule
                         & \multicolumn{4}{c|}{\textbf{MT}}                                & \multicolumn{3}{c}{\textbf{ZSL}} \\        
        \textbf{\# of frames} & 8M              & 16M           & 150M          & N/A  & 16M           & 150M          & N/A  \\ \midrule
        A3C {[}1{]}      & -               & -             & -             & $0.83$ & -             & -             & $0.73$ \\ 
        A3C (Reprod.) & $0.10 \pm 0.01$ & $0.09 \pm 0.04$ & $0.73 \pm 0.01$ & -    & -             & $0.71 \pm 0.02$ & -    \\ \midrule
        DQN              & $0.4 \pm 0.2$     & $0.73 \pm 0.08$ & -             & -    & $0.75 \pm 0.05$ & -             & -    \\
        ACTRCE    & $\textbf{0.69} \pm \textbf{0.04}$   & $\textbf{0.83} \pm \textbf{0.02}$ & -             & -    & $\textbf{0.77} \pm \textbf{0.02}$ & -             & -    \\ 
        \bottomrule
        \end{tabular}
        \vspace{3pt}
    	\caption{The table shows the averaged success rates in multitask and zero-shot scenario for the model trained with DQN versus DQN with \ourmethod{}, compared to A3C published and reproduced result, on ViZDoom single target 5 objects hard mode. The standard deviations are calculated across 2 seeds. MT was evaluated throughout the training process, while ZSL was evaluated at the end of training.}
    	\label{tab:vizdoom_results_5obj_single}
    \end{table*}
    
    \paragraph{Compositional tasks.}
    In KGW, we carried out the experiments in a grid with 3 goals and 3 lavas. A comparison of average success rates over all goals of \ourmethod{}, and baseline DQN is shown in in Fig. \ref{fig:Kgw_result} (c). The shaded area represents the standard deviation over 2 random seeds. We observed that the baseline still could not learn the task while \ourmethod{} learned efficiently and achieved good performance in this highly challenging environment. 

    In ViZDoom, we chose to use 5 objects in \textit{easy} mode for the compositional task, where all 5 objects appeared in front of the agent at the beginning of the episode. Figure \ref{fig:Kgw_result} (g) shows the average success rate when using \ourmethod{} vs. baseline DQN. Likewise, our agent was able to learn very well with hindsight advice, whereas the baseline DQN failed to learn.

    \subsubsection{How much advice do we need?}
    
    We consider a variant of \ourmethod{} where the teacher provides hindsight advice only in the first $\{10\%, 1\%\}$ of frames during training, and stops giving any advice for the remaining portion of the training. 
    We perform this experiment in the ViZDoom environment single target mode with 7 objects, using the GRU as the sentence embedding. 
    Figure \ref{fig:Kgw_result} illustrates the training average success rate over the frames, and Table \ref{tab:vizdoom_anneal} lists the final MT and ZSL performance on the trained agents. 
    We observe that the agent is still able to learn comparably well even given very little (1\%) advice, indicating of the method's robustness in practical settings. 

    \section{Related Work}
	Several approaches have used natural language in the context of reinforcement learning. 
    In pioneering work,~\cite{Maclin94} translated language advice into a short program, implemented as a neural network. This advice network encouraged the policy to take the suggested action. Similarly,~\cite{Kuhlmann04} exploited natural language advice for a RL agent learning to play a (simulated) soccer game. In~\cite{LingNIPS2017}, human feedback in the form of natural language was exploited to shape the reward of an image captioning RL agent. 
	~\cite{BeatAtari} introduced an agent that learns to use English instructions to self-monitor its progress to beat Atari games. This was accomplished by implementing a multi-modal embedding of the pairs of game frames and instructions to determine if the instruction is satisfied, then provide additional reward to the agent. 
	~\cite{LearnLatentLang} proposed to use language as the latent parameter space for few-shot learning problems, including policy search. 
	
	The reverse has also been studied: using reinforcement learning to learn grounded language, referred to as task-oriented language grounding. 
	\cite{misra2017mapping} mapped language instructions and visual observations to actions for manipulating blocks on a 2D plane. 
	\cite{yu2018interactive} grounded language in 2D maze environment, via sentence-directed navigation and question answering task. They introduced the notion of extrapolation zero-shot sentence, where new words are transferred from other use cases and models. In their work, the unseen word in the navigation sentence was seen (transferred from) during the question answering task. In contrast, we transfer the unseen synonym words from a pre-trained sentence embedding InferLite model.
	\cite{bahdanau2018learning} proposed a framework, AGILE, to learn the reward model given the instruction-state pairs from expert trajectories, and train a policy to perform instructions by maximizing the modeled reward. 
	While our work assumes that we are given the teacher to provide the hindsight goal and reward, we can view AGILE as learning to model the teacher's reward function, which is a discriminator outputs state-goal pair $r_\theta(s, g) \in \{0, 1\}$. To fully implement the teacher in ACTRCE, we would also require a generator to generate a goal instruction given the state.
	\cite{co-reyes2018metalearning} considered the case where the language instruction is not fixed for the episode, but can be interactively used to correct policies, in a 2D grid environment. ~\cite{Deepmind3DLang} presented an agent that learns to execute written instructions in a 3D environment through reinforcement learning and unsupervised learning. Our work builds on \cite{chaplot2017gated} who proposed a gated-attention architecture for combining the language and image features to learn a policy that execute written instruction in a 3D environment. 
	
\section{Conclusion}
In this work, we propose the ACTRCE method that uses natural language as a goal representation for hindsight advice. The main goal of the paper is to show that using language as goal representations can bring many benefits, when combined with hindsight advice.
We analyzed the differences among goal representation, and show that ACTRCE can efficiently solve difficult reinforcement learning problems in challenging 3D navigation tasks, whereas HER with non-language goal representation failed to learn. We also show that with language goal representations, the agent can generalize to unseen instructions. With pre-trained language component, the agent can even generalize to instructions with \textit{unseen lexicons}, demonstrating its potential to deal with noisy natural language advice from humans. Although ACTRCE algorithm crucially relies on hindsight advice, we showed that little amount of advice is sufficient for the agent to learn, showing its great practicality.
\subsubsection*{Acknowledgments}

HC was supported by an NSERC CGS-M award. YW is supported by the Google PhD fellowship in machine learning, and an NSERC CGS-D award. We thank Silviu Pitis, Tingwu Wang, Jesse Bettencourt, and Sheldon Huang for helpful feedback on the draft. 

\bibliographystyle{icml2019}
\bibliography{main}

\newpage
\onecolumn
\appendix
\title{Supplementary materials for: }

\section{KrazyGrid World Language Goals} 
\label{app:KGW_goals}
\subsection{More Details on the Environment} 
KrazyGrid World is a 2D grid environment. For our experiments we chose to use 4 functionality of tiles: Goal, Lava, Normal, and Agent. We added an additional colour attribute ranging in 3 colours: Red, Blue, Green. There are 4 discrete actions, ``up", ``down", ``right", ``left". The desired goals in this environment is to reach goals of different colours. We use a global view of the grid state as the observation to the agent.  Each tile in the grid is represented by a concatenation of its functionality and colour attribute, both in one hot vectors. The grid state also includes a one hot vector that represents the agent's position. In our experiments, we use an environment of grid size 9 $\times$ 9, with 3 goals in the environment, each has a distinct colour. We tried various number of lavas in the environment, and made sure there is at least 1 lava of each colour. In the simple task setting, the episode terminates when the agent reaches a lava or a goal, or the episode runs over a maximum time step $T=25$. We automatically generates descriptions give any states of the environment as the teacher. 

 \paragraph{Compositional} We enlarged the goal space by considering the task of compositions of goals. As the goal space changed, we also had to make several modifications to the environment. We let the episode terminate when the agent reaches a lava or reaches 2 different goals, or runs over the maximum time step of $50$. Since we still included simple goals (e.g., ``Reach blue goal") in the goal space, we added an extra action called ``flag". This action can be used for the agent to terminate the episode when it thinks the given goal is accomplished. The environment will be terminated if the agent indeed has accomplished the given task, and will stay unchanged otherwise.

\subsection{List of Language Instructions}
    
\paragraph{Single goal setting} The entire goal space $\mathcal{G}$ consists of 8 goals: $\{$Reach red goal, Reach blue goal, Reach green goal, Reach red lava, Reach blue lava, Reach green lava, Avoid any lava, Avoid any goal$\}$. The desired goal space $\mathcal{G}_d$ consists of 3 goals $\{$Reach red goal, Reach blue goal, Reach green goal$\}$.

\paragraph{Compositional goal setting} The desired goal space $\mathcal{G}_d$ consists of 21 goals $\{$Reach red goal, Reach blue goal, Reach green goal, Reach red goal and Reach blue goal, Reach blue goal and Reach red goal, Reach red goal or Reach blue goal, Reach blue goal or Reach red goal, Reach red goal and Reach green goal, Reach green goal and Reach red goal, Reach red goal or Reach green goal, Reach green goal or Reach red goal, Reach blue goal and Reach green goal, Reach green goal and Reach blue goal, Reach blue goal or Reach green goal, Reach green goal or Reach blue goal, Reach red goal and Avoid any lava, Avoid any lava and Reach red goal, Reach blue goal and Avoid any lava, Avoid any lava and Reach blue goal, Reach green goal and Avoid any lava, Avoid any lava and Reach green goal$\}$. 

The entire goal space $\mathcal{G}$ has another of 17 goals: $\{$Reach red lava or Reach blue lava, Reach blue lava or Reach red lava, Reach red lava or Reach green lava, Reach green lava or Reach red lava, Reach blue lava or Reach green lava, Reach green lava or Reach blue lava, Reach red lava, Reach red lava and Avoid any goal, Avoid any goal and Reach red lava, Reach blue lava, Reach blue lava and Avoid any goal, Avoid any goal and Reach blue lava, Reach green lava, Reach green lava and Avoid any goal, Avoid any goal and Reach green lava, Avoid any lava and Avoid any goal, Avoid any goal and Avoid any lava$\}$. 

\section{ViZDoom Environment Detail} \label{app:vizdoom}
\subsection{Environment Description}

\textbf{ViZDoom}\citep{Kempka2016ViZDoom,chaplot2017gated}: The 3D learning environment is based on the first person shooter game Doom. 
	At each time step, the state is a raw pixel image from first person perspective of the 3D environment.
	The environment consists of a room containing several random doom objects of various types, colours, and sizes. 
	The agent can navigate the environment via 3 possible actions: turn left, turn right, and move forward. 
	The goal for the episode is a natural language instruction in the form "Go to the [\textit{target attribute}] [\textit{object}]", such as "\textit{Go to the green torch}". 
	Only one target object is present in each episode.
	The episode terminates either when the agent reaches an object (regardless of whether it is correct or not), or the maximum time step is reached ($T=30$). 
	A reward of 1 is given if the correct object is reached, otherwise a reward of 0 is given.
	The environment has several difficulty modes, depending on how the objects and the agent are distributed at the beginning of the episode.
	In the \textit{easy} mode, the agent is spawned at a fixed location facing forward.
	The objects are placed evenly spaced apart along a horizontal  row in front of the agent.
	In \textit{hard}, both the objects and the agent are randomly spawned, and the objects are not necessarily in view.
	We focus on the easy and hard mode for our experiments.
    \vspace{-0.05cm}
    
\subsection{ViZDoom Composition Instructions}
The compositional instructions consist of two single object instructions from the original training set joined by the word "and", such as "\textit{Go to the red torch and Go to the pillar}". We did not include any superlative instructions, such as "\textit{Go to the largest object}".

\begin{figure*}[t!]
    \begin{center}
    \begin{tabular}{p{0.4\textwidth}p{0.4\textwidth}}
    \includegraphics[width=0.4\textwidth]{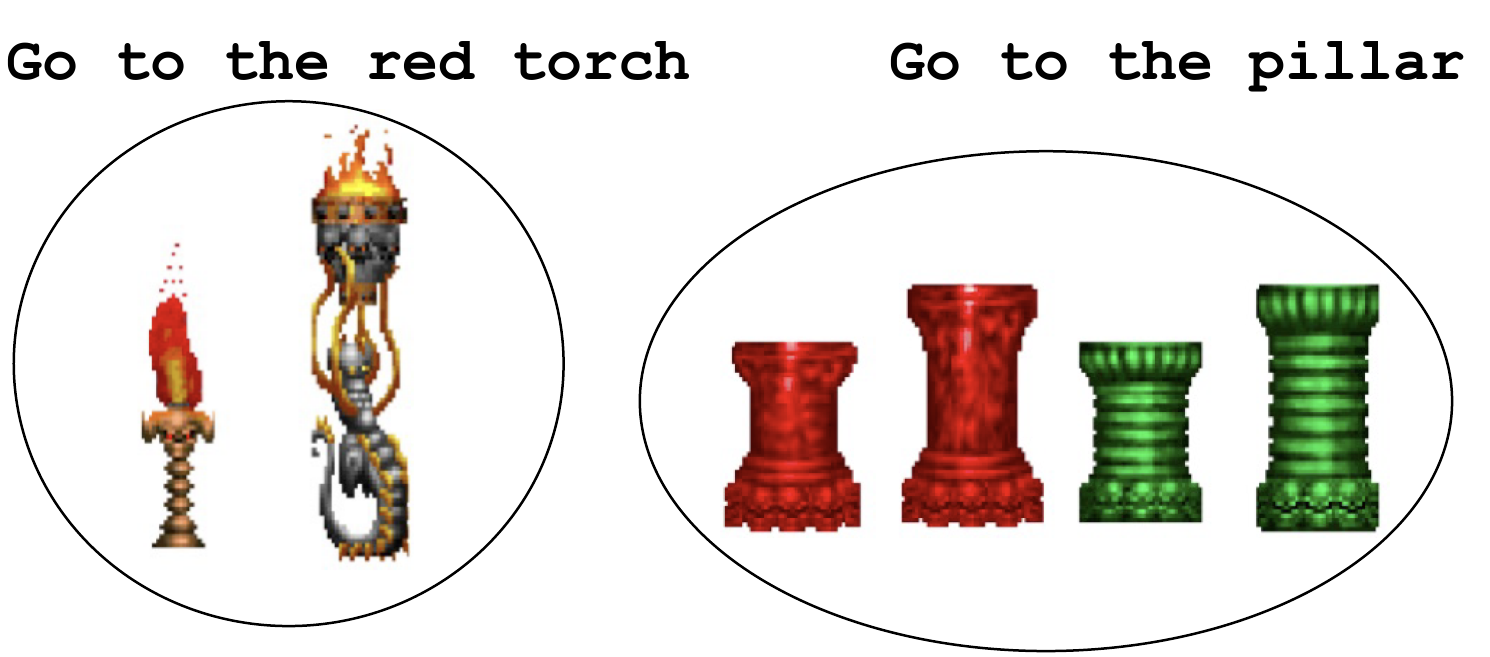} &
    \includegraphics[width=0.4\textwidth]{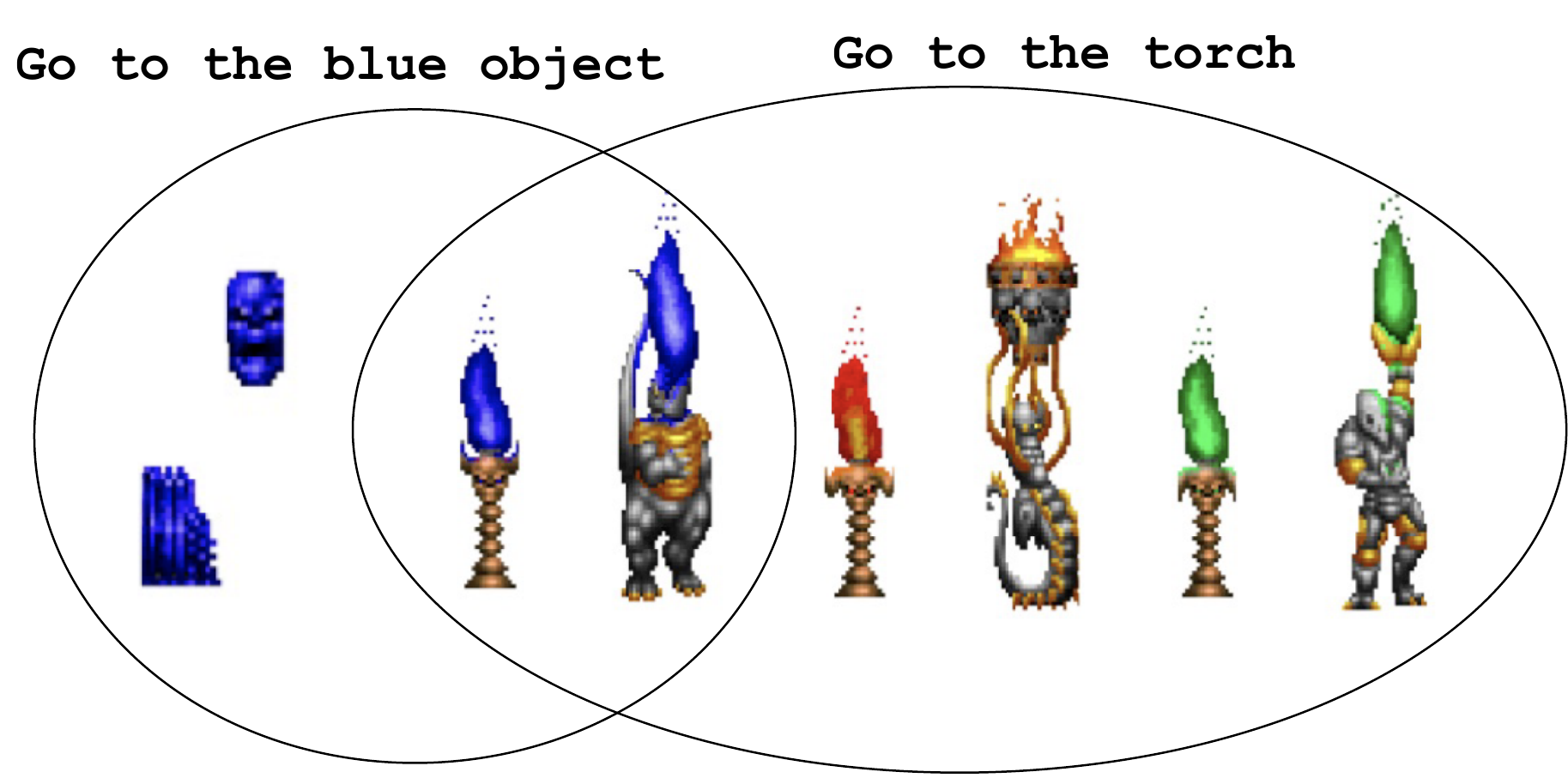}\\
    \centering(a)&\centering(b)
    \end{tabular}
    \caption{Example of (a) unambiguous composition of instructions and (b) ambiguous composition of instructions}
    \label{fig:vizdoom_comp}
    \end{center}
\end{figure*}

We ensured that the desirable instructions were unambiguous--given two instructions, the set of objects satisfying the first instruction is mutually exclusive from the set of objects satisfied in the second instruction. For example, "Go to the torch and Go to the pillar" have mutually exclusive set of valid objects. An ambiguous compositional instruction may have objects that satisfy both instructions. For example, "Go to the blue object and Go to the torch" is ambiguous because a blue torch satisfies both instructions. This is illustrated in Figure \ref{fig:vizdoom_comp}.

\subsection{ViZDoom Composition: Modification for State}
\begin{figure*}[t!]
    \begin{center}
    \begin{tabular}{p{0.3\textwidth}p{0.3\textwidth}p{0.3\textwidth}}
    \includegraphics[width=0.3\textwidth]{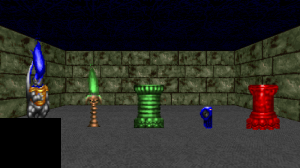} &
    \includegraphics[width=0.3\textwidth]{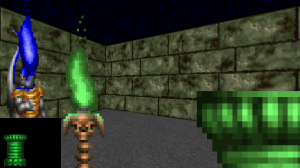} &
    \includegraphics[width=0.3\textwidth]{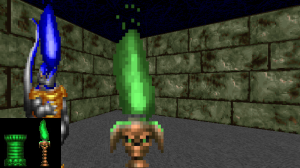}\\
    \centering(a)&\centering(b)&\centering(c)
    \end{tabular}
    \caption{Example state input screens seen by the agent in a ViZDoom composition task episode. The composition instruction for this episode is "\textit{Go to the green short torch and Go to the green short pillar}". (a) Shows the starting state. (b) is the state when the agent has reached the green short pillar. (c) is the final state once the agent reached the short green torch.}
    \label{fig:vizdoom_comp_traj}
    \end{center}
\end{figure*}

For the ViZDoom composition task, we modify the raw pixel image of the environment to include a basic head-up display (HUD) consisting of black rectangle in the bottom left of the screen. When the agent reaches an object, a small thumbnail image of the reached object will appear inside the HUD. The HUD can show up to 2 objects reached, and the episode terminates once the again has reached a second object. Figure \ref{fig:vizdoom_comp_traj} illustrates what the agent sees as the input image throughout a composition task episode.

\subsection{ViZDoom Synonym Instructions}
\label{sec:synonyms}
We generate the synonym instructions by replacing an original word in the instruction with one of the synonyms listed in Table \ref{tab:vizdoom_syn_list}.

\begin{table}[h]
    \centering
    \begin{tabular}{ll}
    \toprule
    \textbf{Original Word}                   & \textbf{Synonym}\\ \hline
        Go to & Reach, Approach, Get to \\
        largest & biggest\\
        smallest & littlest \\
        small & tiny \\
        large & big, huge \\
        short & low, little \\
        tall & big, towering \\
        red & crimson \\
        blue & indigo, teal \\
        green & olive, lime \\
        yellow & amber, gold, sunny \\
        torch & lamp, beacon, lantern, flaming object \\
        pillar & column, pedestal \\
        skull & cranium, head \\
        key & opener, unlocker \\
        card & badge, pass \\
        armor & shield \\
        object & thing, item \\
    \bottomrule
    \end{tabular}
    \caption{Original vocabulary and corresponding synonyms}
    \label{tab:vizdoom_syn_list}
\end{table}

\subsection{ViZDoom Teacher Advice Generation}
\paragraph{\textbf{Singleton task}.} When the agent does reach an object (correct or incorrect), we generate a positive teacher advice by sampling from the training set of instructions that can describe the reached object.  Additionally, we also generate a negative feedback by randomly picking one of the  object in the current environment that was not reached, then sampling an instruction that described that unreached object, while ensuring that it will not describe the reached object. 

 When the agent did not reach any object, we simply do not give any positive teacher advice (i.e. the achieved goal). We had originally tried to give the advice of “\textit{Reach no object}” when the agent did not reach any objects at the end of the episode, but it did not improve the performance. However, we still give the unachieved goal, from any of the objects in the current environment.
 
 \paragraph{\textbf{Compositional task}.} We follow a similar process for compositional task, depending on the number of objects reached.  If the agent reached 0 or 1 object only, we generate a singleton advice in similar fashion as singleton task. We later found that not giving any advice when reaching 0/1 object also performed equally well and slightly faster learning. 
 If the agent reached 2 objects during the episode, for positive advice we sample generate singleton instructions for each of the two objects, ensuring that they were not the same instructions, then merge them with "\textit{and}" keyword. For negative advice, we generate singleton instructions for the unreached objects and combine two of them, ensuring that the instructions do not refer to the reached objects at all. 
    
\section{Architectures} \label{app:arch}
\paragraph{KrazyGrid World}
\begin{itemize}
    \item \textbf{Singleton} We use series of convolution layers follows with ReLU activation functions for prepossessing the grid observation: 32 of 3x3 kernel with stride 1, followed by 64 of 3x3 kernel size with stride of 2 and then 64 of 3x3 kernel with a stride of 1 and 128 of 3x3 kernel with a stride of 2. We input the language sentences as word level one hot vectors to a LSTM of hidden size 128. The LSTM's last hidden vector is passed into a fully connected layer with 128 output units with sigmoid activation. This acts as the attention vector that is multiplied channel-wise with the 64 feature maps from the convolutional later. The gated feature maps are then flattened and passed through a fully connected layer with ReLU activation with 256 units. We then pass into a linear layer to predict the 4 action values.  
    \item \textbf{Compositional} We use series of convolution layers follows with ReLU activation functions for prepossessing the grid observation: 32 of 1x1 kernel with stride 1, followed by 32 of 3x1 kernel size with stride of 2. We input the language sentences as word level one hot vectors to a bidirection LSTM, of hidden size 40. After obtaining the preprocessed observation, we augment with the history vector provided by the environment as a query to attend to all the hidden states in the Bi-direction LSTMs via Luong attention \cite{Luong2015EffectiveAT}. We then keep processing the
    observation using two more convolution layers follows with ReLU activation functions: 64 of 3x3 kernel with stride 2, followed by 64 of 3x3 kernel size with stride of 2. At each of these two layers, we use the context vector formed by language attention to attend back the observation by gating on 64 feature maps in a channel-wise fashion. The final gated feature maps are then flattened and passed through a fully connected layer with ReLU activation with 256 units. We then pass into a linear layer to predict the 5 action values.  
\end{itemize}
\paragraph{ViZDooms} Our architecture is almost identical to \cite{chaplot2017gated}, except that we uses a linear output layer for the action values, and we did not use dueling architecture. The state input to the network is RGB image of size $3\times300\times168$. A series of convolution layers follows with ReLU activation functions: 128 of 8x8 kernel with stride 4, followed by 64 of 4x4 kernel size with stride of 2 and then 64 of 4x4 kernel with a stride of 2. To process the language input, we first use an embedding matrix to embed the voculary to a vector of size 32. We use the Gated Recurrent Unit (GRU) \cite{GRU} of size 256 as the recurrent cell to process word embeddings. The GRU's last hidden vector is passed into a fully connected layer with 64 output units with sigmoid activation. This acts as the attention vector that is multiplied channel-wise with the 64 feature maps from the convolutional later. The gated feature maps are then flattened and passed through a fully connected layer with ReLU activation with 256 units. We then pass into the LSTM with 256 hidden units, and its hidden units go through a linear layer to predict the 3 action values. 
	
\section{Training Details} \label{app:training}

    \paragraph{KrazyGrid World} For KrazyGrid World experiments, we tune the following hyperparameters within corresponding range: learning rate $\{0.0003,0.001,0.003\}$, replay buffer size of $\{5000, 10000\}$, training the network every $\{1, 2, 4\}$ frames. We generate episodes with $\epsilon$-greedy policy starting at $\epsilon=1.0$ then decaying linearly to $0.01$ by $10000$ frames and remain at $0.01$.
    We use Double DQN \cite{DoubleDQN} to reduce the Q-value overestimation and Huber loss ($\delta=1$) for stable gradients. 
    
    \paragraph{ViZDoom}For ViZDoom environment, we use the same set of 56 training instructions to train the agent, and a set of 15 held out test instructions for zero-shot evaluation as in \cite{chaplot2017gated}. 
    We also used their code \cite{DeepRLGrounding} for reproducing training using Asynchronous Advantage Actor Critic (A3C) \cite{A3C}.
    We implemented our version of DQN on top of the implementation of Arnold \cite{arnold, arnoldGit}.
    The cyclic buffer replay buffer contains the last 10000 and 20000 most recent transitions for the easy and hard mode, respectively, chosen from the range $\{1000, 10000, 100000\}$ and fine tuned.
    We generate episodes with $\epsilon$-greedy policy starting at $\epsilon=1.0$ then decaying linearly to $0.01$ by $10000$ frames and remain at $0.01$.
    We found that this performed better than using a larger $epsilon=0.1$.
    Only after the first 1000 frames have been collected that the training begins, selected from a range of $\{1000, 10000, 100000\}$.
    We use Double DQN \cite{DoubleDQN} to reduce the Q-value overestimation and Huber loss ($\delta=1$) for stable gradients. 
    We use the Adam optimizer \cite{kingma2014adam} with a learning rate of 0.0001.
    The network is updated every 4 frames on easy and 16 frames on difficult mode.
    While updating less often reduces sample efficiency, in practice we found that this speeds up training wall clock time and converges to a better performance.
	When sampling from the replay buffer, we sample the start a random episode and select 32 consecutive frames from the replay buffer.
	This approach leads to more accurate estimates of the hidden state of the LSTM.
	In addition, the sequential mini-batch allows us to perform $n$-step Q learning as outlined in \cite{A3C}. 
	The correlation between samples in the mini-batch is alleviated by running 16 parallel threads to send gradient updates to the shared network parameters.
    We synchronize the training thread model with the shared network each time before computing the training loss and gradient. 
    The target network is synchronized with the current model once every 500 time steps. 
    and use one additional thread to evaluate the multi-task success rate throughout the training process.
    
\section{Different types of teachers}
\label{sec:teacher_types}
In this section we described the details of teacher types. Firstly, note that not all of the language descriptions describe favorable behaviors. For example, in the KGW, a desirable goal is ``reach a goal", but there are also undesirable states that corresponds to goal ``reach a lava". Hence we consider a subset $\mathcal{G}_{d}\subseteq\mathcal{G}$ to denote all desired goals that the agent is expected to perform. At each episode, we first sample a goal $g\in\mathcal{G}_d$ from the set of all desired language goal spaces, and ask the agent to explore the environments conditioned on the goal. When the episode ends, we obtained an advice from a teacher $\mathbb{T}$ by observing the terminal state, $g^*=\mathbb{T}(s_T)$. Depending on the kind of teacher, we obtain a different kind of advice or no advice. We consider three kinds of teacher as follows:
	\begin{itemize}
	    \item \textbf{Optimistic} A teacher who gives advice only when the agent achieved a desirable goal, i.e., when $g^*\in\mathcal{G}_d$. When the agent performs poorly, there is no advice from this teacher.
	    \item \textbf{Knowledgeable} A teacher who describes what the agent has achieved in all scenarios, including the undesirable behaviors, as advice to the agent.
	    \item \textbf{Discouraging} A teacher who describes a desired goal $g^*\in\mathcal{G}_d$ that the agent has not achieved as advice to the agent. In this case, the trajectory with goal replaced by the teacher's advice will receive a reward of 0.
	\end{itemize}

\subsection{How does each teacher perform?}
	In this section, we investigated how different teachers help in giving advice. We compared our method to the DQN algorithm. We denote the method using only optimistic and discouraging teachers as ``\ourmethod{}$^-$". We denote the method using \emph{knowledgeable} teachers as well as discouraging teachers as \ourmethod{} (this is the version we use in the main paper). We evaluated three methods in KrazyGrid World and the results are as follows.
    
   We tried two different kind of grids, one with 3 lavas and the other with 6 lavas, and both with 3 goals of different colours. Fig. \ref{fig:Kgw_result} (a) and (b) show the average success rate over all goals on 16-environments training. The shaded area represents the standard deviation over 2 random seeds. The baseline DQN is shown in blue curve, which failed to learn anything at all. ``\ourmethod{}$^-$" (shown in orange) quickly learned and achieved good results on both environments. However, we observed that when the number of lavas increased, the task became harder, and \ourmethod{}$^-$ performed worse after 32 millions frames of training (the performance dropped from $83\%$ to $63\%$).  On the other hand, we observed that having knowledgeable teachers always helped speed up learning. In particular, when the number of lavas increased, the task became harder for \ourmethod{}$^{-}$. However, with knowledgeable teachers, language advice was provided even when the agent reached lava, and hence the amount of advice per step was kept the same in the more difficult setting. Therefore, we observed \ourmethod{} learning at a similar rate in the more difficult setting, leaving a big gap from \ourmethod{}$^{-}$  after 32 millions frames of training ($80\%$ vs $63\%$).

\begin{figure*}[h!]
\begin{center}
\begin{tabular}{p{0.3\textwidth}p{0.3\textwidth}p{0.3\textwidth}}
\hspace{-0.2cm}\includegraphics[width=0.33\textwidth]{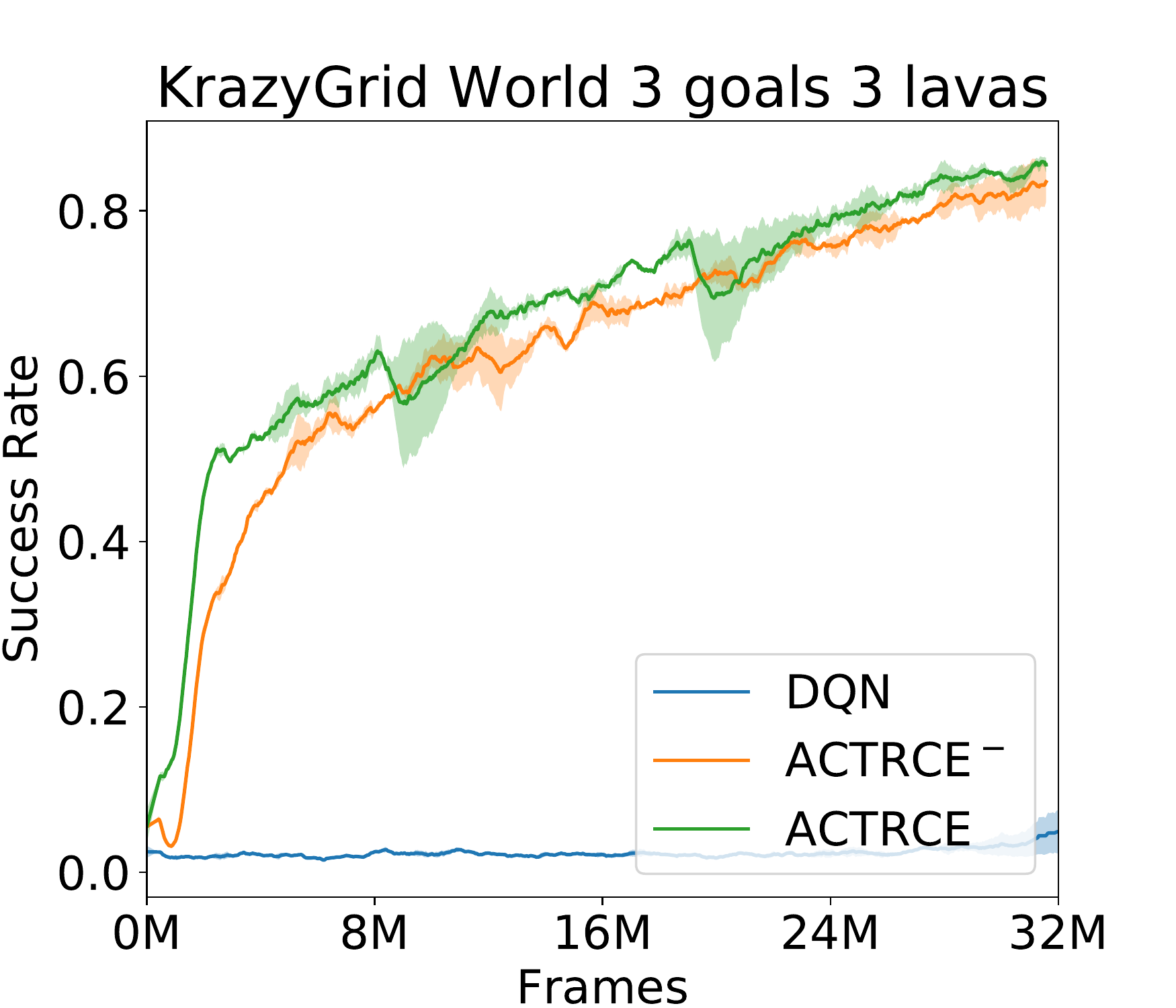}&
\hspace{-0.2cm}\includegraphics[width=0.33\textwidth]{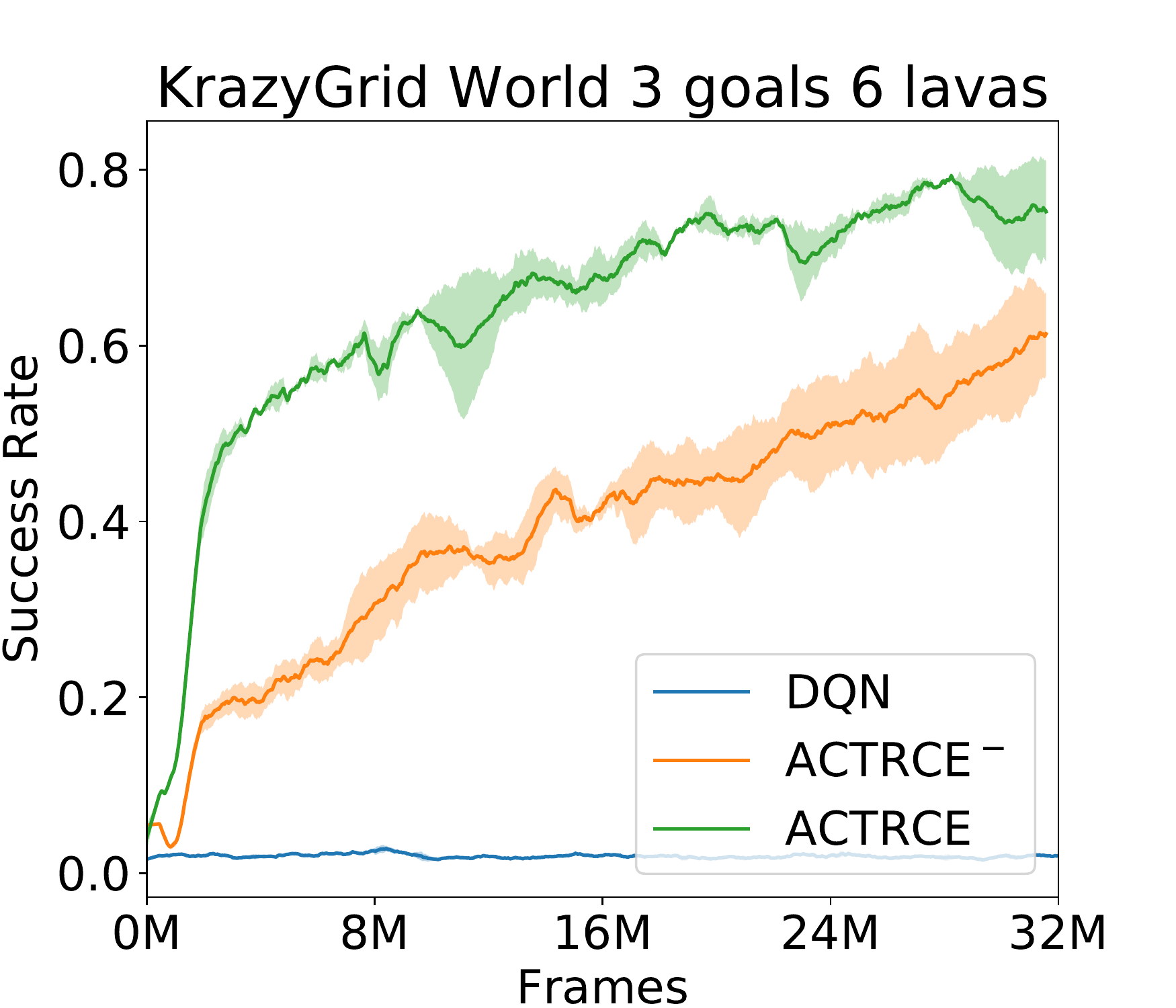}&
\hspace{-0.2cm}\includegraphics[width=0.33\textwidth]{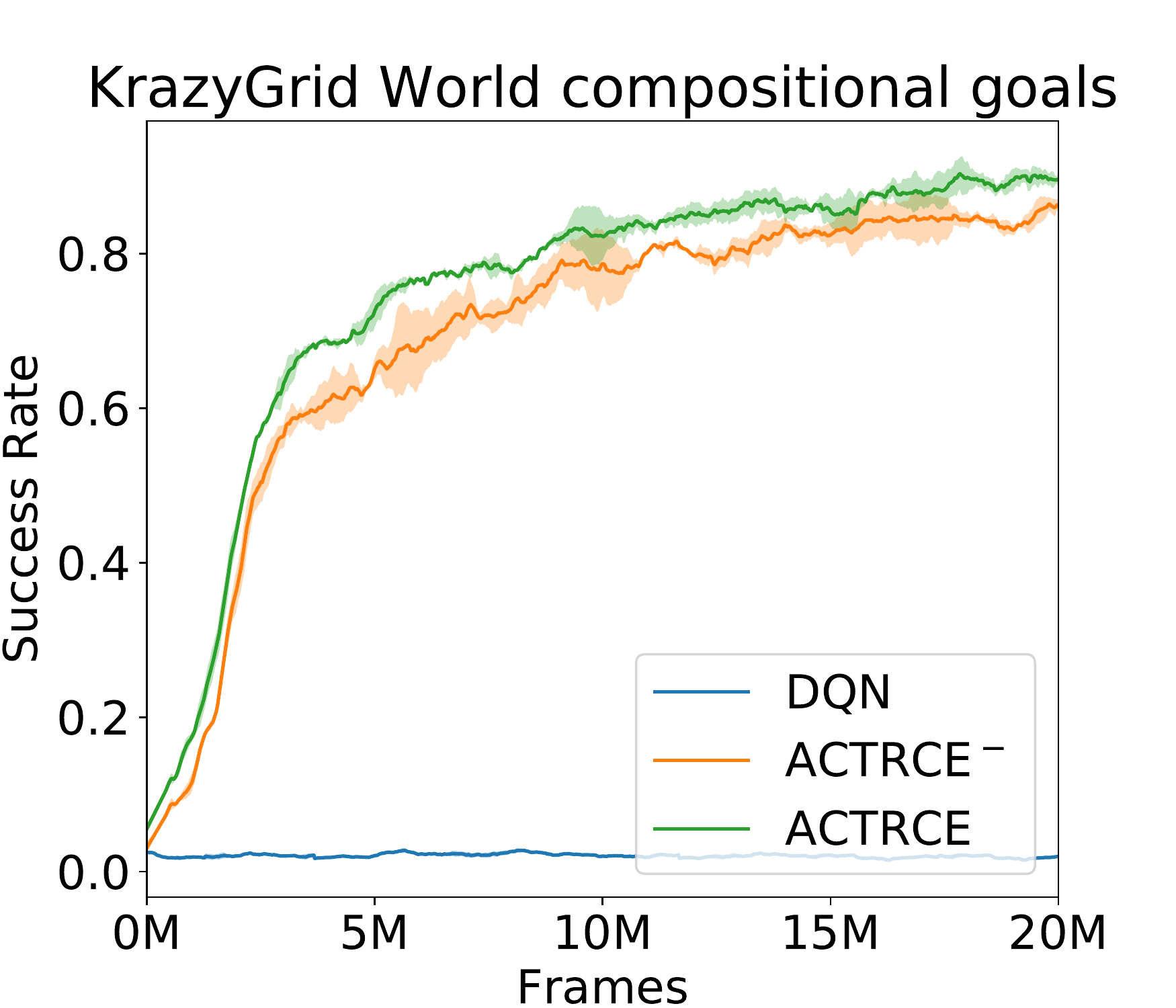}\\
\centering(a)&\centering(b)&\centering(c)
\end{tabular}
\caption{Performance comparisons on KrazyGrid World environments. The success rates are calculated over all desired goals and 16 different environments. Shaded area represents standard deviation 2 random seeds.}
\label{fig:Kgw_result_appendix}
\end{center}
\end{figure*}

\section{Can learning easy goals help learning difficult goals?}

As we mentioned, there are desirable tasks as well as undesirable tasks. One would worry what if the desirable tasks are very difficult, would the agent only learns to accomplish easier tasks which are undesirable? Hence, we would like to ask whether learning the easier tasks from hindsight can provide any learning signal for more difficult tasks. We hypothesize that it is possible in the KGW setting as learning to ``reach lava" can aid learning controllers, which is used to ``reach goals". 

\begin{figure}[h!]
\begin{center}
\begin{tabular}{p{0.35\textwidth}p{0.35\textwidth}}
\includegraphics[width=0.35\textwidth]{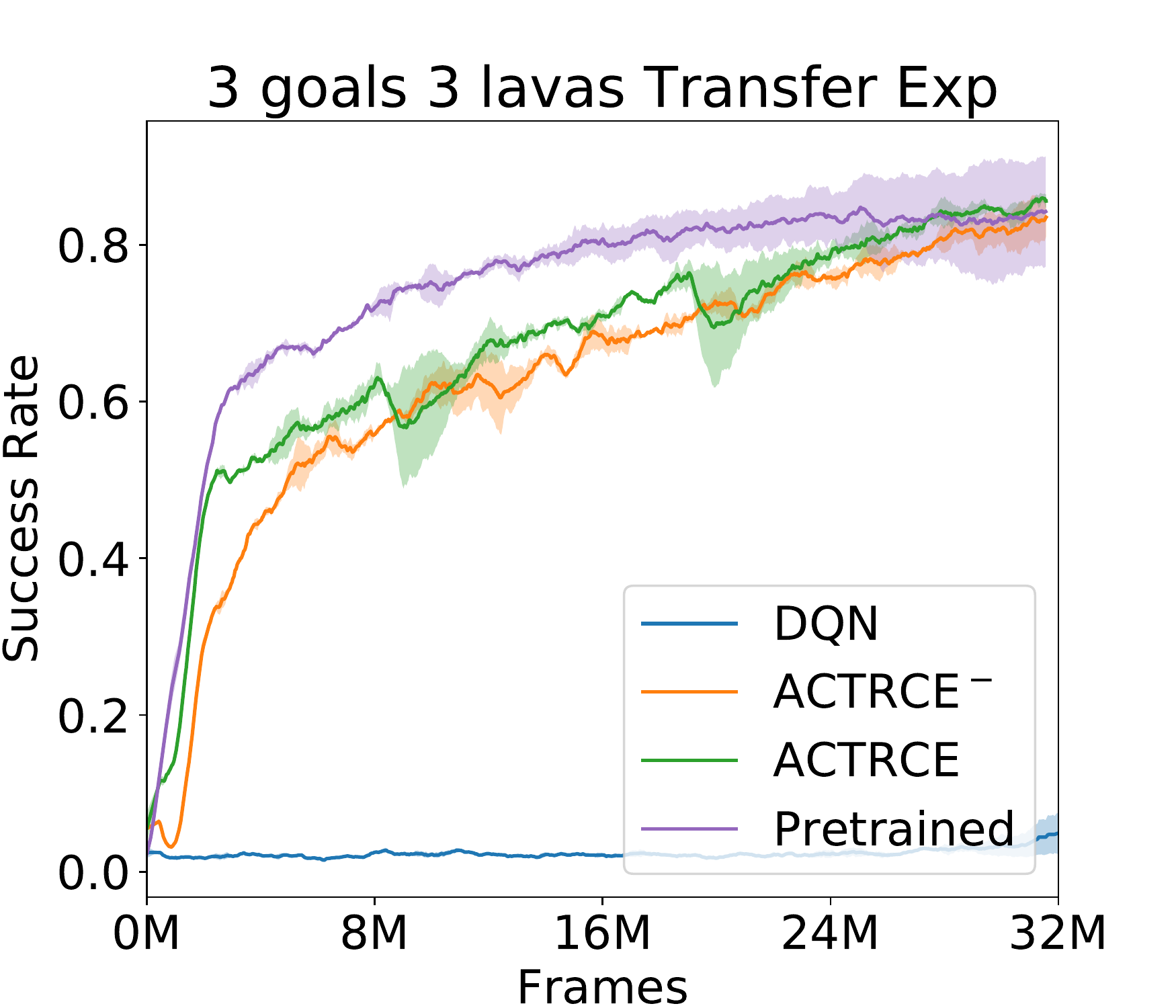}&
\includegraphics[width=0.35\textwidth]{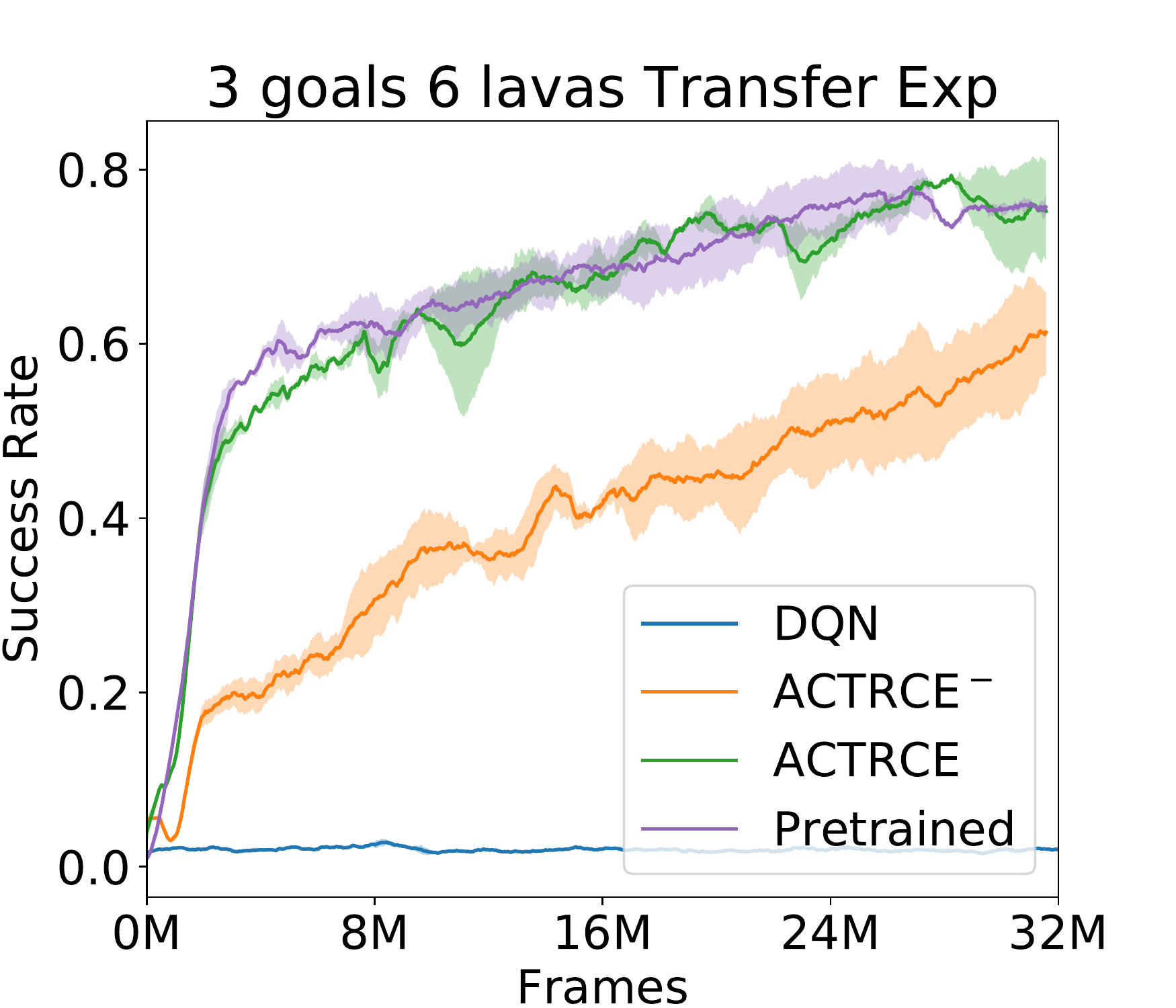}
\end{tabular}
\caption{\small Transfer experiments that demonstrate pre-training with easier goals can aid learning more difficult goals.}
\label{fig:transfer}
\end{center}
\end{figure}

We designed the following transfer learning experiment to further our hypothesis. First, we artificially constructed a \textbf{pessimistic} teacher who only gives advice when the agent achieved undesirable goals, i.e., when $g^*\in\mathcal{G}\textbackslash\mathcal{G}_d$.
 We pretrained the agents with only the pessimistic teacher for 5 million frames. Now, we carried out training using \ourmethod{}$^-$ (with optimistic and discouraging teachers) with the pretrained agent and compared to unpretrained ones. Results are shown in Figure \ref{fig:transfer}. We found that by pretraining the agent using pessimistic agents, even though the language goals in the pretraining phase have no overlapping with the actual training goals, the agent learned much faster than the unpretrained ones. In particular, in the environment with 3 lavas, the pretrained agent learned the fastest. In the environment with 6 lavas, pretrained agents learned as fast as \ourmethod{}, leaving a big gap from \ourmethod{}$^-$, even though it was given the same amount of advice during training as \ourmethod{}$^-$. This provides an evidence that learning easier goals can sometimes provide learning signals for harder goals, especially when both tasks require similar modules.

\section{ViZDoom Average Episode Length} \label{app:vizdoom_epslen}
\begin{figure*}[h!]
    \begin{center}
    \begin{tabular}{p{0.32\textwidth}p{0.32\textwidth}p{0.32\textwidth}}
    \hspace{-0.5cm}\includegraphics[width=0.32\textwidth]{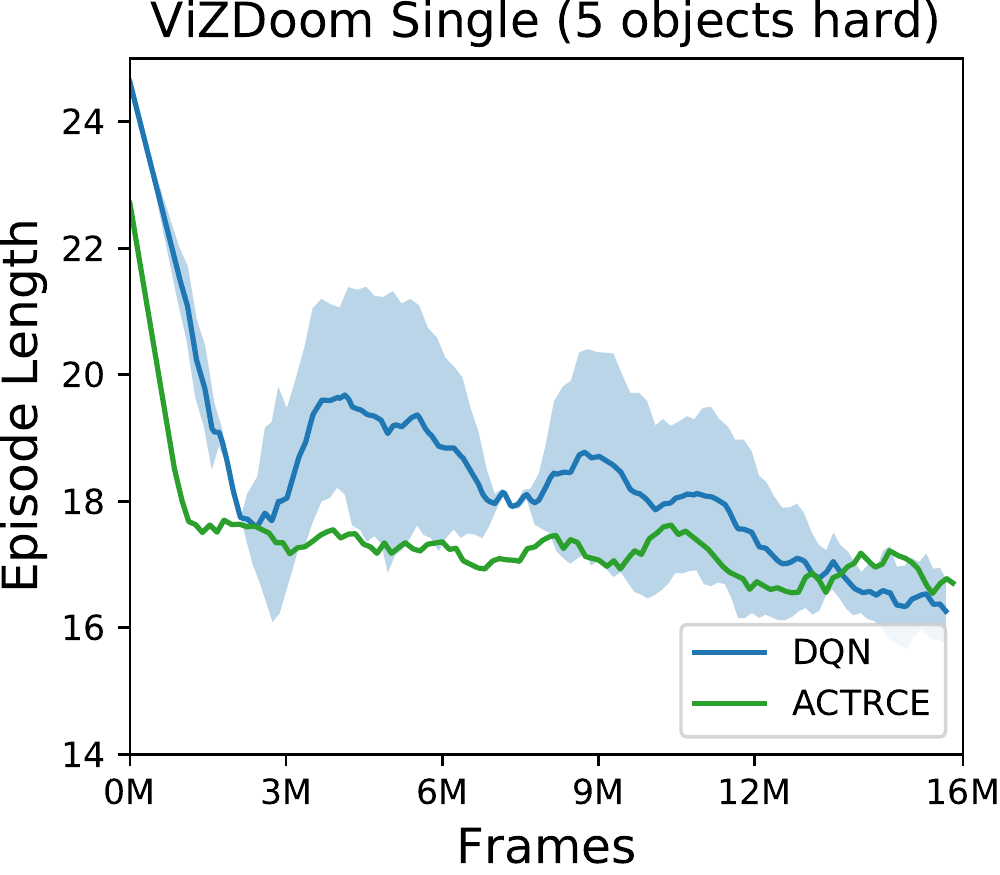} &
    \hspace{-0.5cm}\includegraphics[width=0.32\textwidth]{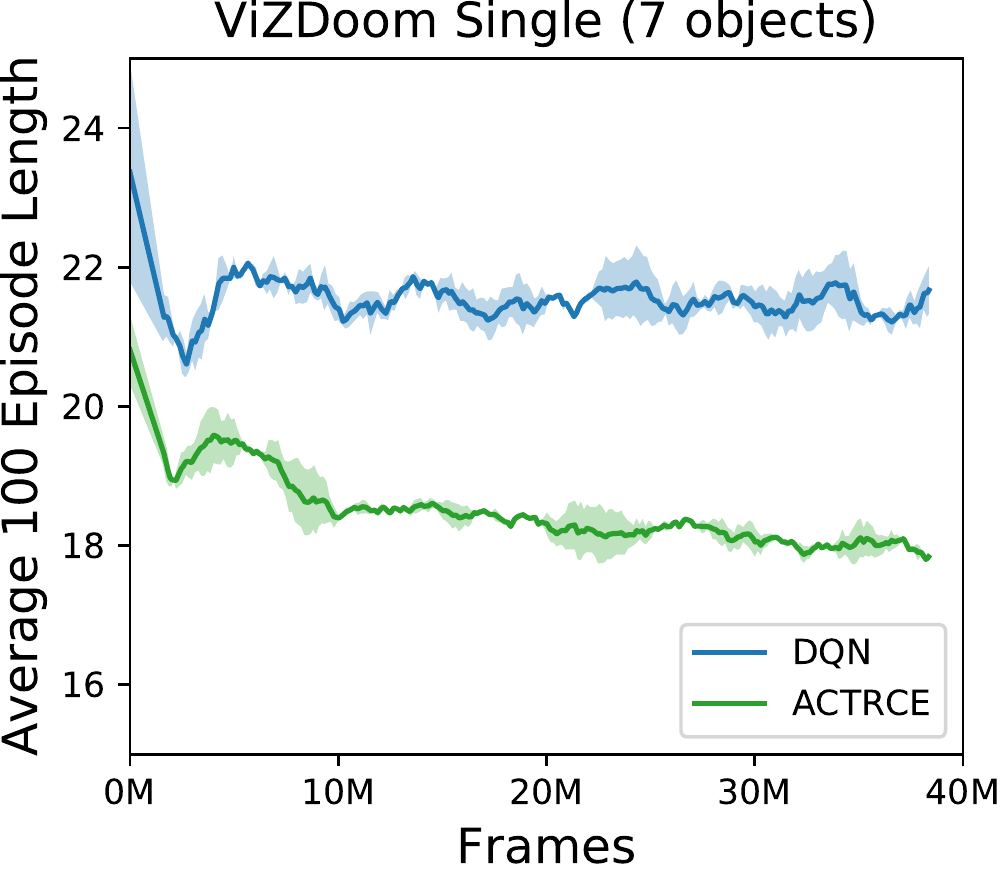} &
    \hspace{-0.5cm}\includegraphics[width=0.32\textwidth]{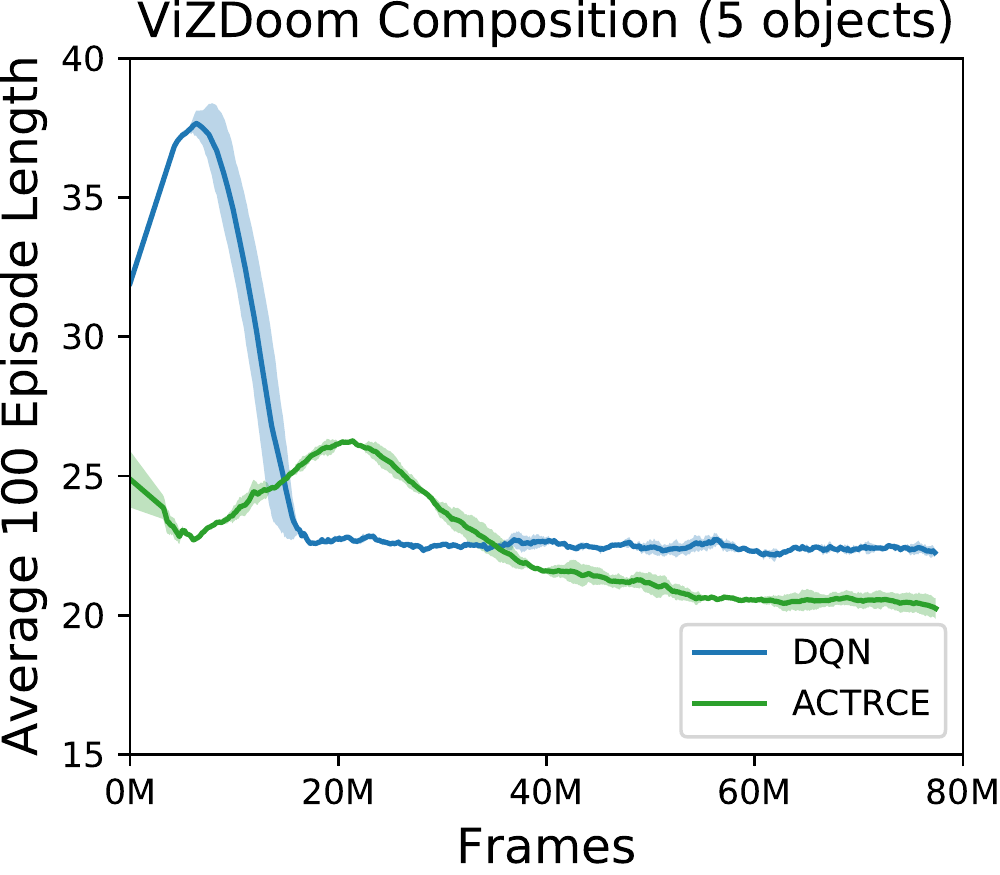}\\
    \hspace{-0.7cm}\centering(a)&\hspace{-0.7cm}\centering(b)&\hspace{-0.7cm}\centering(c)
    \end{tabular}
    \caption{Average training episode lengths in VizDoom environments. (a) 5 objects in hard mode, in single target case. (b) 7 objects in hard mode, in single target case. (c) 5 objects in easy mode, in composition case.}
    \label{fig:vizdoom_epslen}
    \end{center}
\end{figure*}

We report on the average episode length (averaged over 100 episodes) during training for 3 ViZDoom scenarios: single target 5 and 7 objects in hard mode, and composition target 5 objects in easy mode. The GRU hidden state representation of the sentence were used in the plots in Figure \ref{fig:vizdoom_epslen}. We observe that the average episode length is slowly decreasing when using ACTRCE, while the baseline DQN remains fairly flat for the harder 7 objects (single target) and 5 objects (compositional).

\section{ViZDoom Cumulative Success Rate vs. Episode Length Curve} \label{app:vizdoom_eps_success}
We take the final trained model and run 100 episodes, noting whether each run was successful or not (binary value). We construct a \textit{cumulative success rate} (CSR) versus episode length curve, where the x-axis is the episode length, and the y-axis is the fraction of total number of episodes that were successful and had episode length \textit{less than or equal} to the x-axis value:
\begin{align}
    \text{CSR}(x) = \frac{\text{\# of successful eps with eps length} \leq x}{\text{Total number of episodes}}    
\end{align}
Therefore, the curve is monotonically increasing, with the y-axis being the overall success rate when the x-axis value is the maximum episode length. The better the model, the larger the area under this curve will be---similar in spirit to the precision-recall curve---because it will be able to have more of successful trajectories that are short early on. 

\begin{figure*}[h!]
    \begin{center}
    \begin{tabular}{p{0.32\textwidth}p{0.32\textwidth}p{0.32\textwidth}}
    \hspace{-0.5cm}\includegraphics[width=0.32\textwidth]{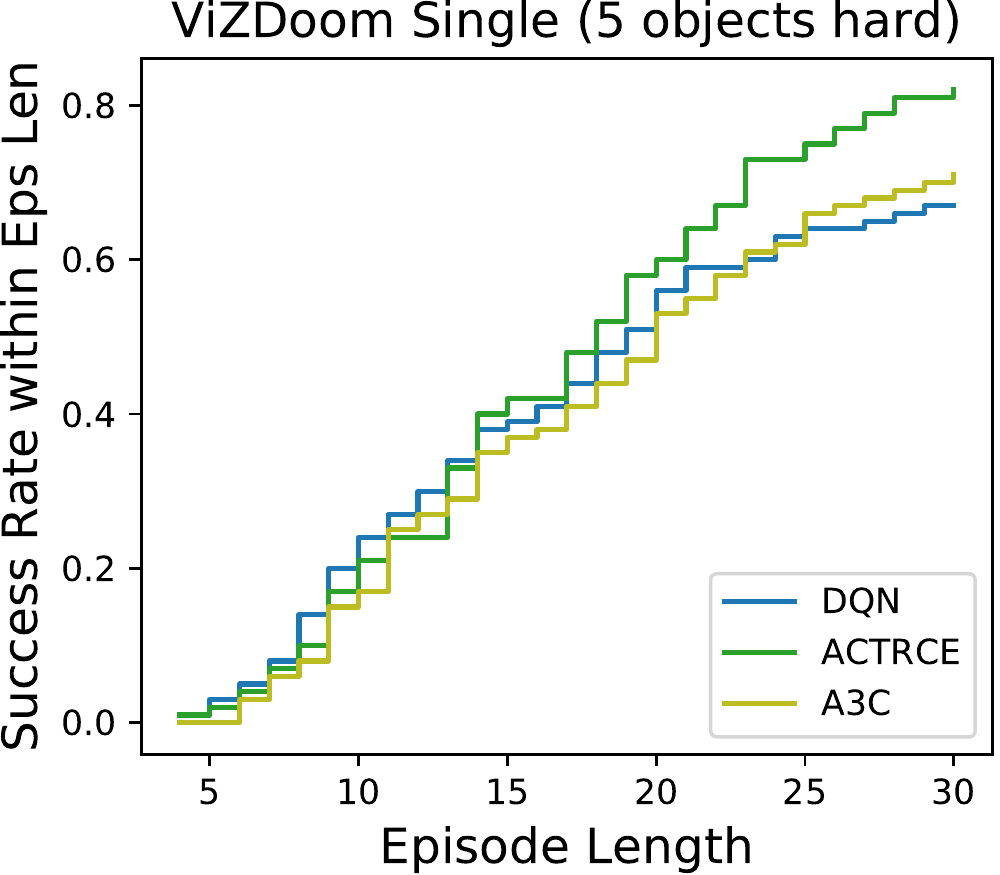} &
    \hspace{-0.5cm}\includegraphics[width=0.32\textwidth]{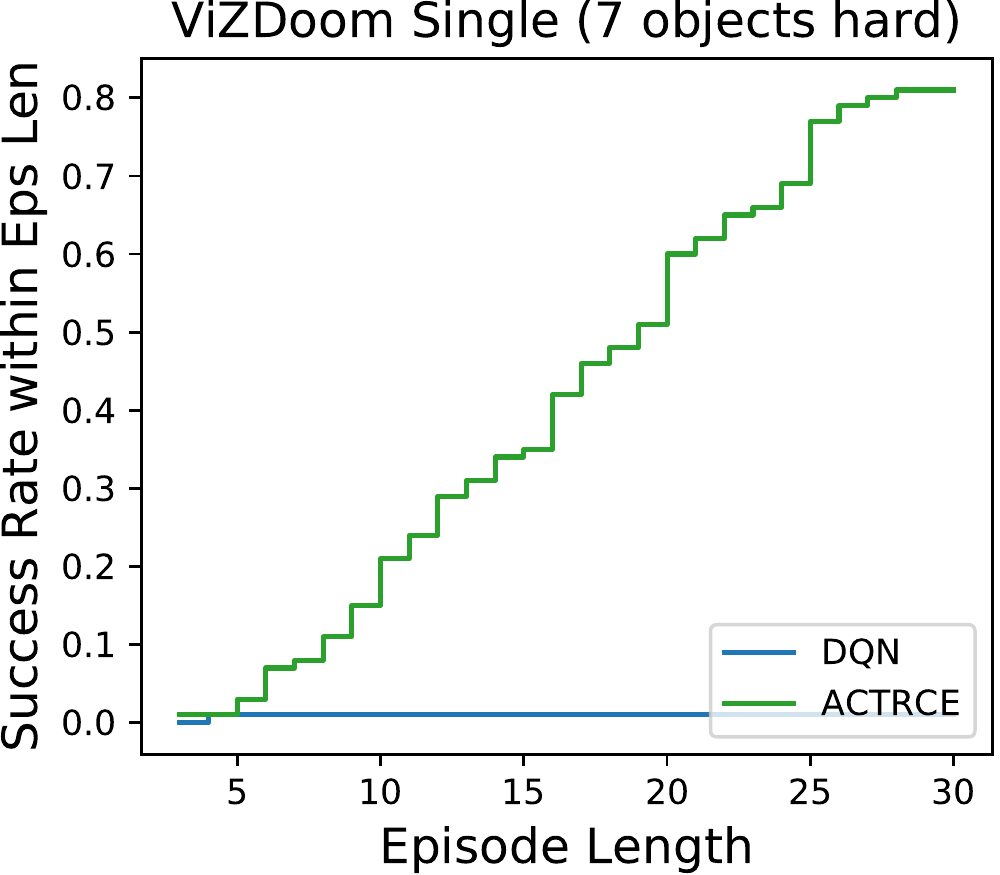} &
    \hspace{-0.5cm}\includegraphics[width=0.32\textwidth]{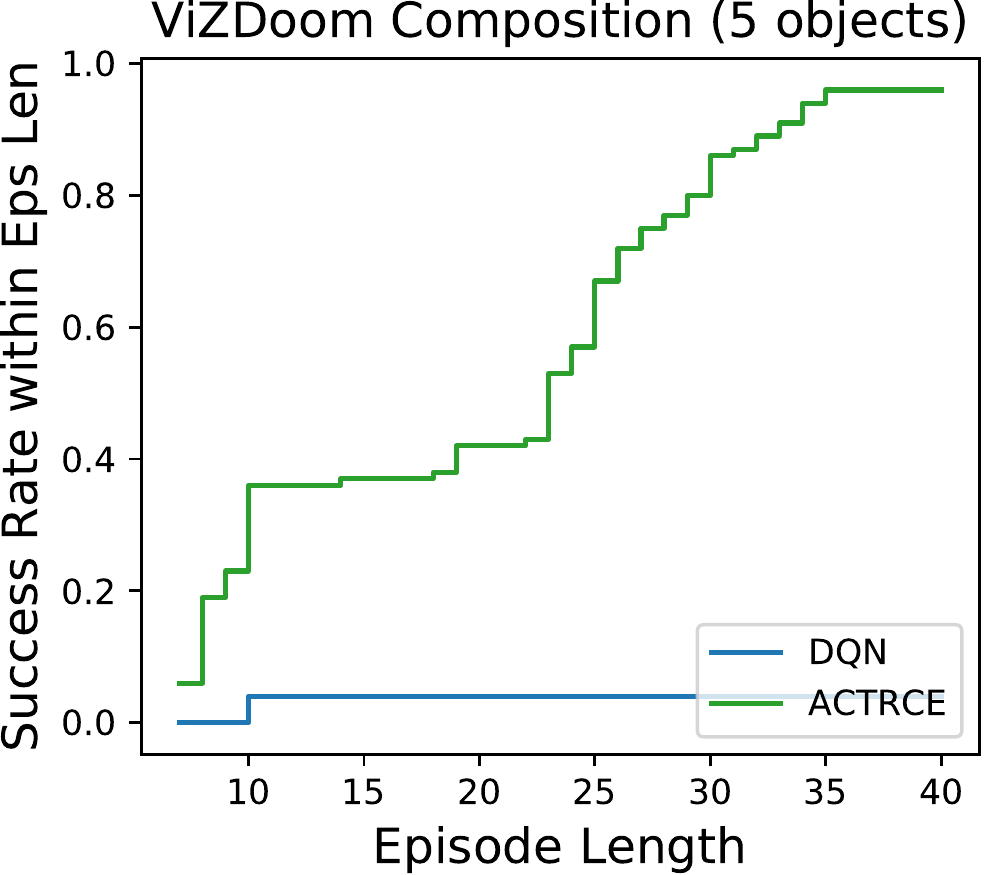}\\
    \hspace{-0.7cm}\centering(a)&\hspace{-0.7cm}\centering(b)&\hspace{-0.7cm}\centering(c)
    \end{tabular}
    \caption{Multi-task (MT) cumulative success rate versus episode in ViZDoom environments, using GRU hidden state sentence representation. (a) 5 objects in hard mode, in single target case. (b) 7 objects in hard mode, in single target case. (c) 5 objects in easy mode, in composition case.}
    \label{fig:vizdoom_epslen_curve}
    \vspace{-0.3cm}
    \end{center}
\end{figure*}

Figure \ref{fig:vizdoom_epslen_curve} shows the Multi-task (MT) cumulative success rate for the 3 ViZDoom environment tasks, using GRU hidden state language encoding: single target 5 and 7 objects in hard mode, and composition target 5 objects in easy mode. In the 5 objects hard mode (Figure \ref{fig:vizdoom_epslen_curve}a), we observe that all 3 training algorithms had similar performance until around episode length of 20, where \ourmethod{} has more successful trajectories which are longer. In the 7 objects hard mode (Figure \ref{fig:vizdoom_epslen_curve}b), \ourmethod{} maintains a similar behaviour while the baseline DQN essentially was only able to get success on very short episodes (i.e. when the target object was very close by). Lastly, in the 5 objects composition task (Figure \ref{fig:vizdoom_epslen_curve}c), the curve for \ourmethod{} indicates that there were 2 groups of trajectories: one requiring less than 10 time-steps, while another requiring over 20 time-steps. The former group occurs when the two target objects are adjacent to one another, making it easy to reach the second object after the first in only a few time steps. The latter group occurs when the target objects are not adjacent to each other, and thus requires the agent to more carefully turn around and avoid hitting other objects when trying to reach the second target. 

\section{Additional details on Hindsight Advice Annealing}
\label{app:anneal_advice}
We report the results for the hindsight advice annealing when using InferLite and One-Hot representation when providing hindsight advice on the first 10\% and 1\% of training: 
\begin{table}[h]
    \centering
    \begin{tabular}{lcc}
    \toprule
    \textbf{Method}                   & \textbf{MT} & \textbf{ZSL} \\ \hline
    ACTRCE (GRU) 1.0   & $\bf{0.80 \pm 0.06}$ & $\bf{0.76 \pm 0.03}$  \\
    ACTRCE (GRU) 0.1   & ${0.75 \pm 0.01}$ & ${0.74 \pm 0.06}$  \\
    ACTRCE (GRU) 0.01   & ${0.73 \pm 0.04}$ & ${0.66 \pm 0.05}$  \\
    \hline
    ACTRCE (InferLite) 1.0   & $\bf{0.80 \pm 0.01}$ & $\bf{0.76 \pm 0.02}$  \\
    ACTRCE (InferLite) 0.1   & ${0.78 \pm 0.04}$ & ${0.68 \pm 0.01}$  \\
    ACTRCE (InferLite) 0.01   & ${0.76 \pm 0.01}$ & ${0.68 \pm 0.01}$  \\
    \hline
    ACTRCE (OneHot) 1.0   & $\bf{0.80 \pm 0.03}$ & $-$  \\
    ACTRCE (OneHot) 0.1   & ${0.74 \pm 0.06}$ & $-$  \\
    ACTRCE (OneHot) 0.01   & ${0.70 \pm 0.06}$ & $-$  \\
    \bottomrule
    \end{tabular}
    \caption{Comparing the Multitask and Zero-shot Generalization when the agent has limited hindsight advice to only the beginning of training.}
    \label{tab:vizdoom_anneal}
    \vspace{3pt}
    \end{table}
    
\begin{figure}[h!]
    \begin{center}
    \begin{tabular}{p{0.35\textwidth}p{0.35\textwidth}}
    \hspace{-0.2cm}\includegraphics[width=0.35\textwidth]{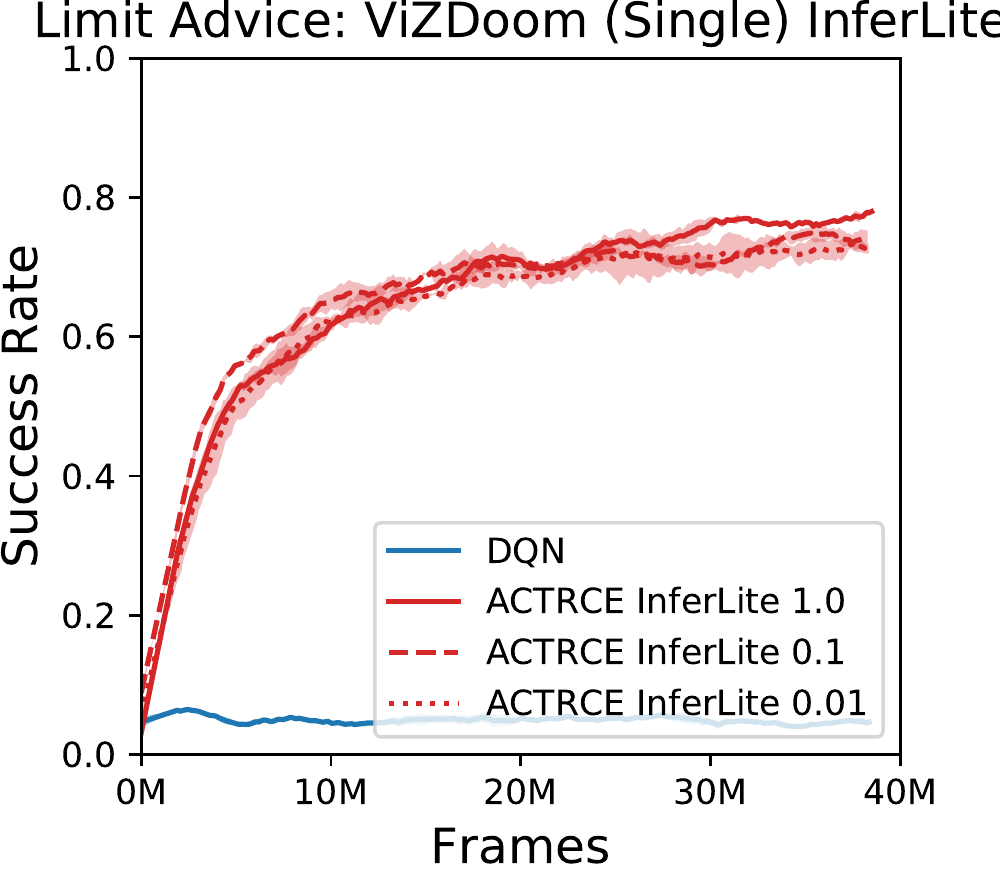} &
    \hspace{-0.2cm}\includegraphics[width=0.35\textwidth]{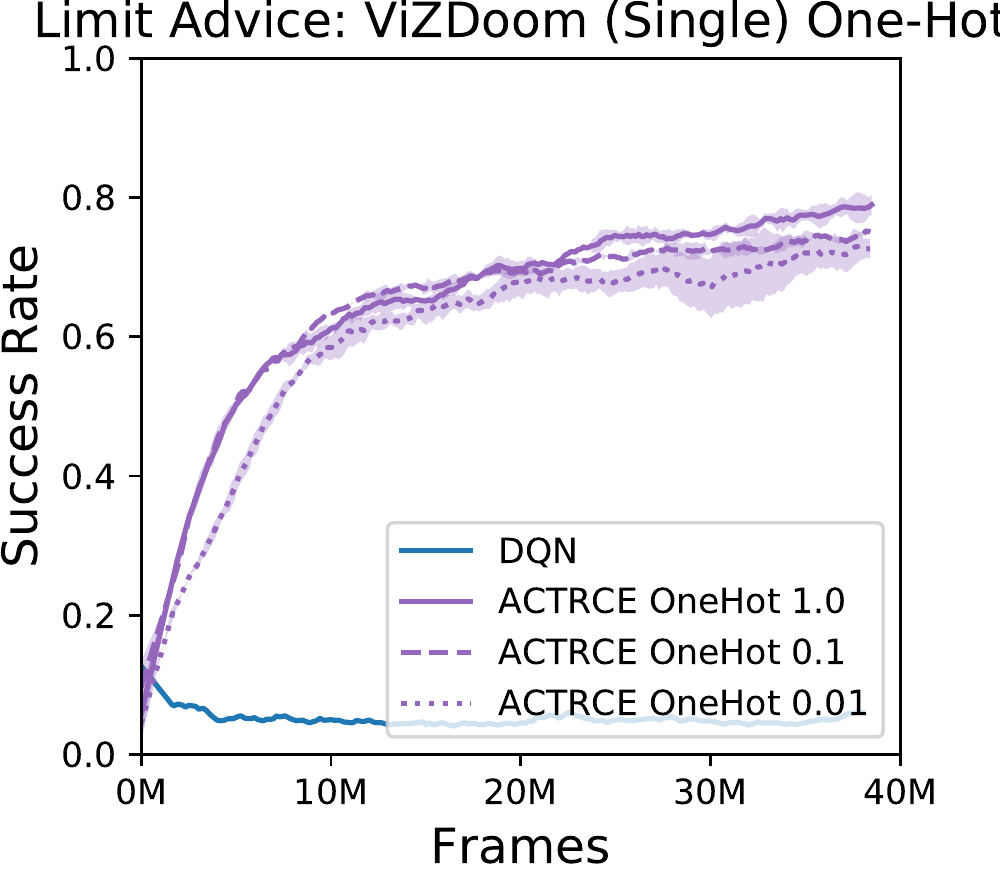} \\
    \centering(a)&\centering(b)
    \end{tabular}
    \caption{Performance comparisons on VizDoom environment.}
    \label{fig:vizdoom_anneal_inferlite_onehot}
    \end{center}
\end{figure}

\section{Visualization} \label{app:visualization}
For each single-target instruction in the training and test set (70 instructions), we obtain an embedding vector of dimension 256 using either GRU, InferLite, or One Hot (storing entire embedding). From this embedding matrix, we can visualize the learned instruction embedding in several ways.

\subsection{Embedding Correlation Comparison} \label{app:visualization_correlation}
For each pair of instructions $i$ and $j$, we obtain their embeddings vectors $v_i$ and $v_j$, and compute their \textit{correlation distance}, $\text{cdist}(v_i, v_j)$. We define the \textit{correlation distance} between two vectors $u$ and $v$ as:
\begin{align}
    \text{cdist}(u, v) = 1 - \frac{(u - \bar{u})\cdot(v - \bar{v})}{||(u - \bar{u})||_2 ||(v - \bar{v})||}
\end{align}

We divide each matrix by the maximum entry as to scale the results to $[0,1]$.

\begin{figure*}[h!]
\begin{center}
	\centering
    \includegraphics[width=1.0\textwidth]{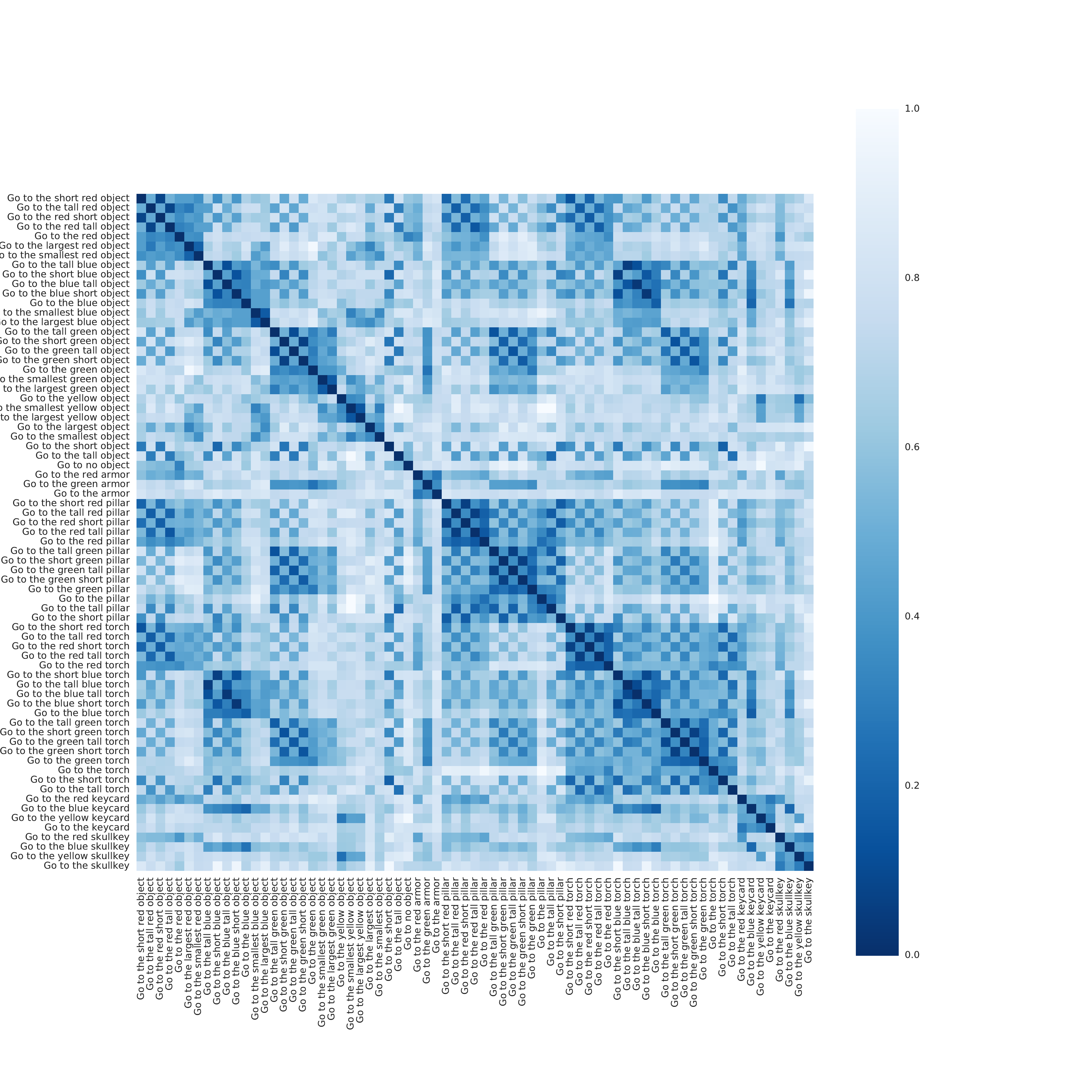}
	\caption{Correlation matrix for GRU. Best seen in electronic form.}
	\label{fig:attn}
	\end{center}
\end{figure*}

\begin{figure*}[h!]
\begin{center}
	\centering
    \includegraphics[width=1.0\textwidth]{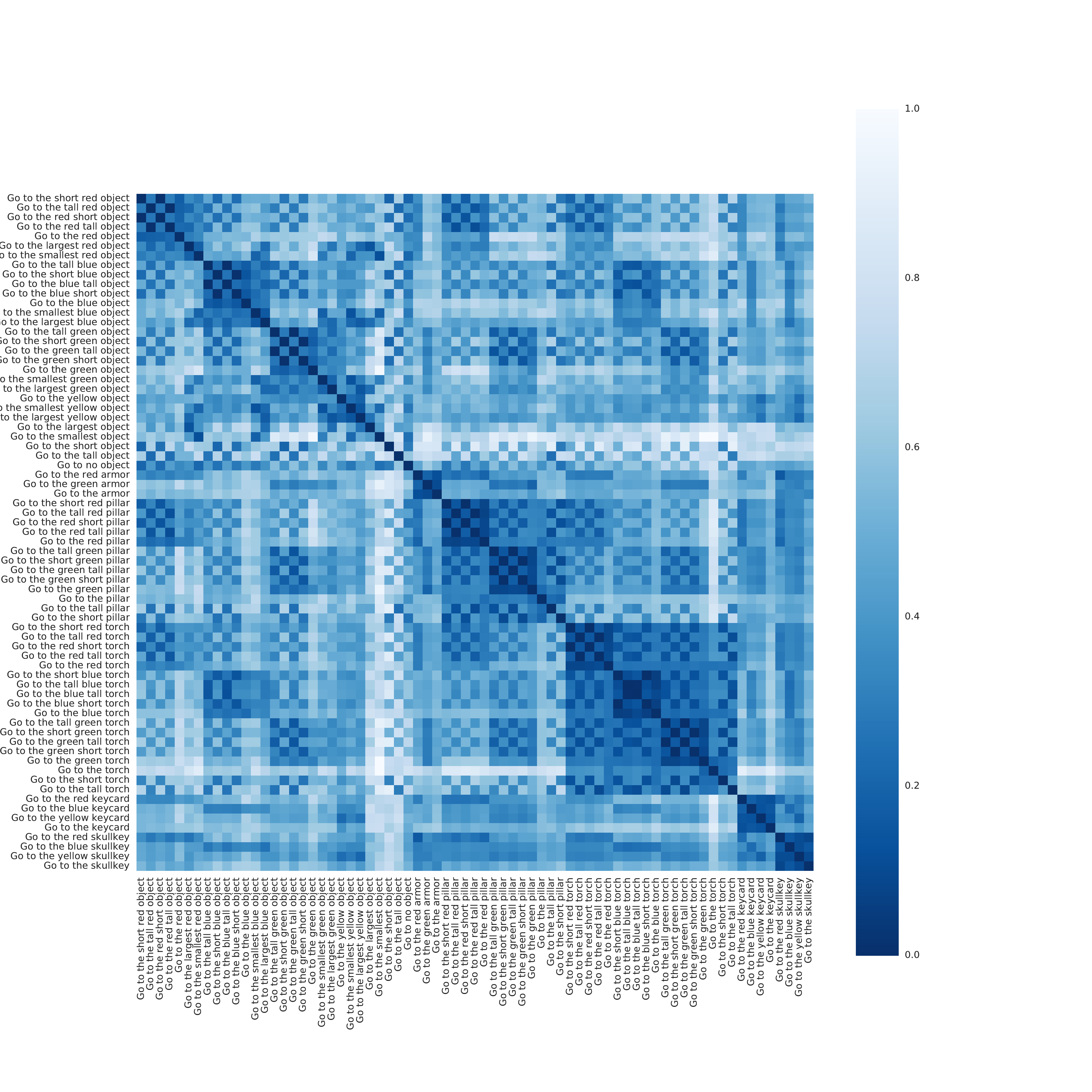}
	\caption{Correlation matrix for InferLite. Best seen in electronic form.}
	\label{fig:attn_interlite}
	\end{center}
\end{figure*}

\begin{figure*}[h!]
\begin{center}
	\centering
    \includegraphics[width=1.0\textwidth]{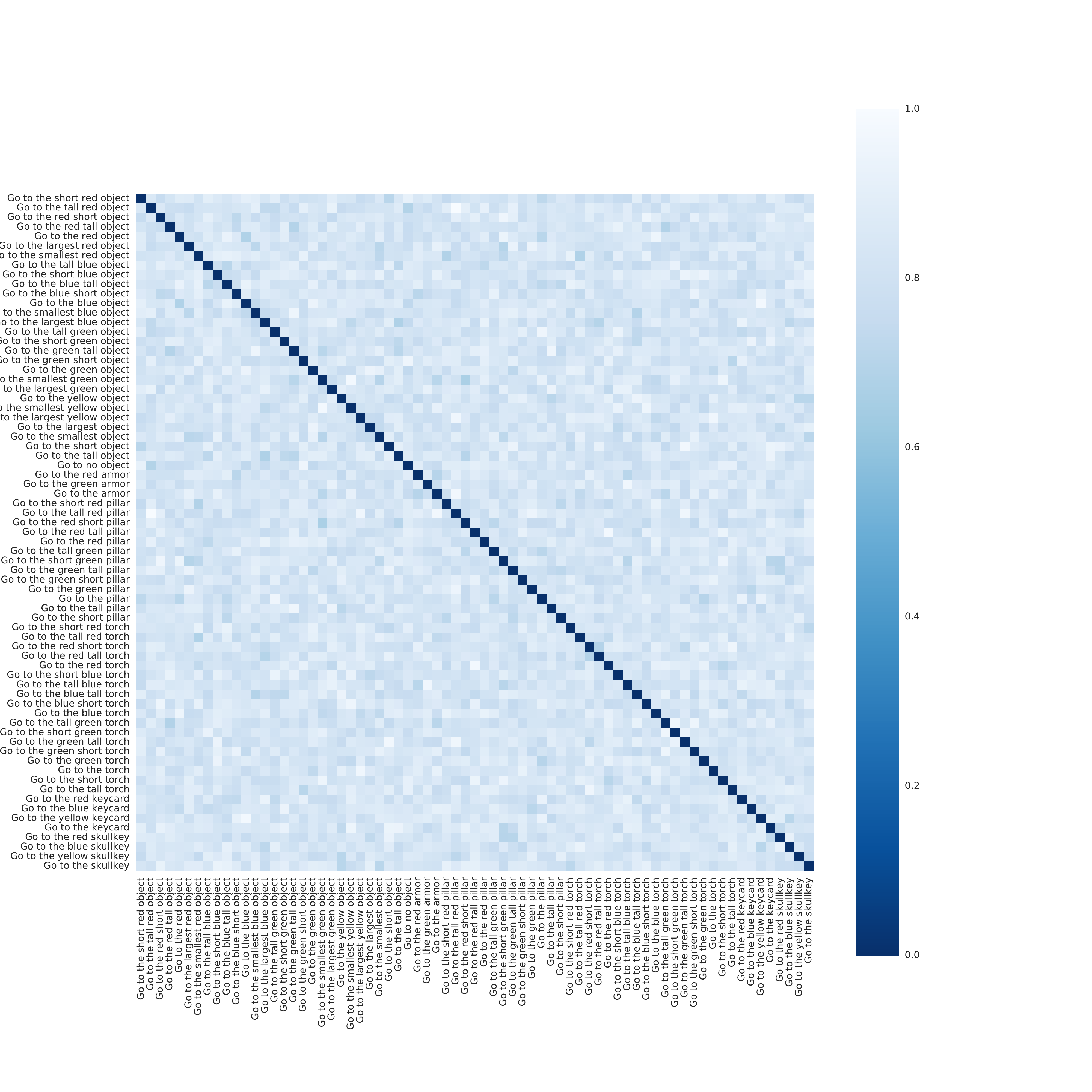}
	\caption{Correlation matrix for One-Hot. Best seen in electronic form.}
	\label{fig:attn_onehot}
	\end{center}
\end{figure*}

\subsection{t-SNE Plot Comparison}
We use t-SNE to visualize the space of instruction embeddings. In the following figure, the shape of the datapoint indicates the type of object, such as triangle for armor, diamond for pillar, etc., and the colour indicate the object's colour, with black being used for when no colour is specified. Finally the size of the data point corresponds to the size indicated in the object, from 'smallest', 'small', none specified, 'large', and 'largest'. For the instructions with synonym word replacement, we simply leave the colour as black and the default size.

\begin{figure*}[h!]
\begin{center}
		\begin{subfigure}{.50\textwidth}
        \includegraphics[width=1.0\textwidth]{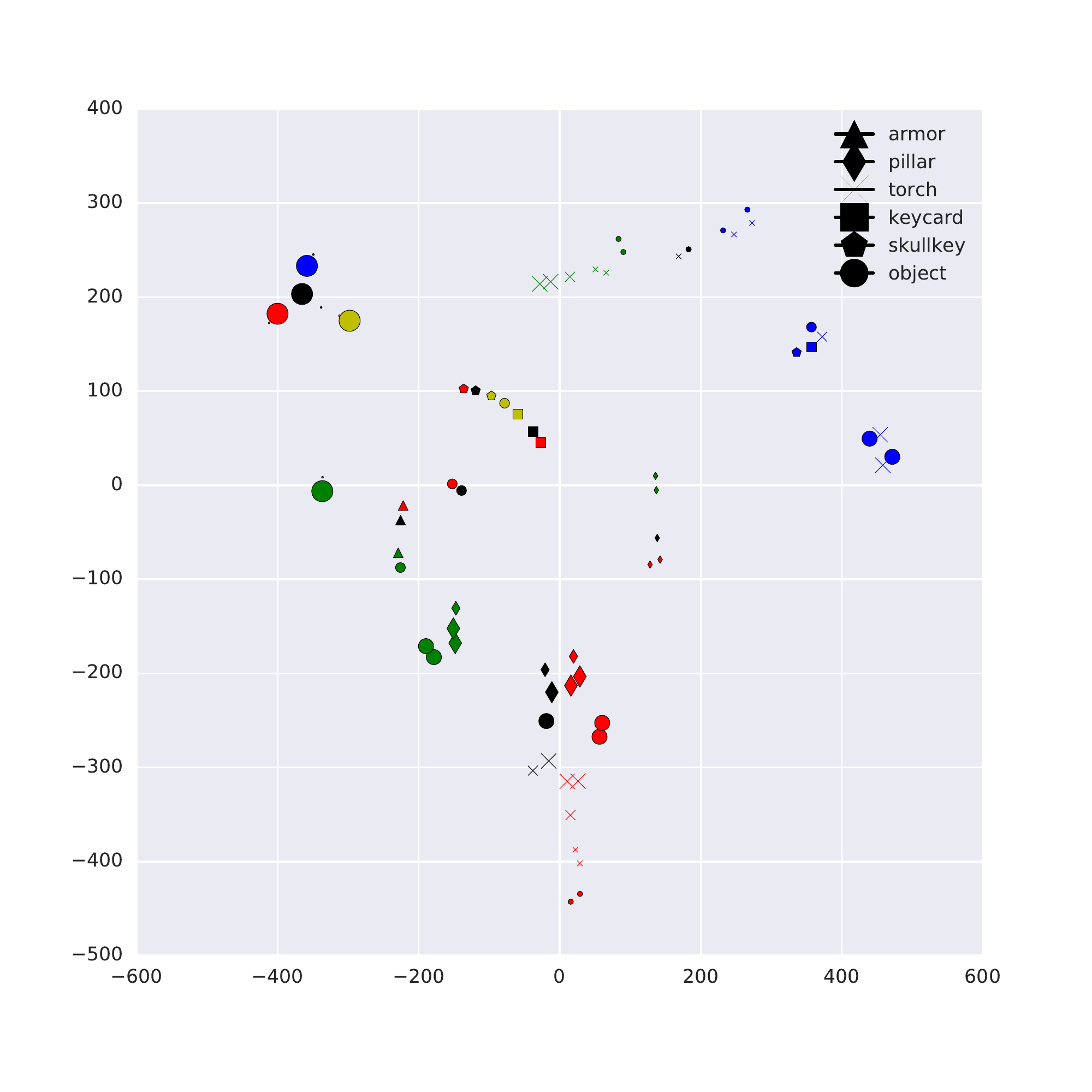}
        \includegraphics[width=1.0\textwidth]{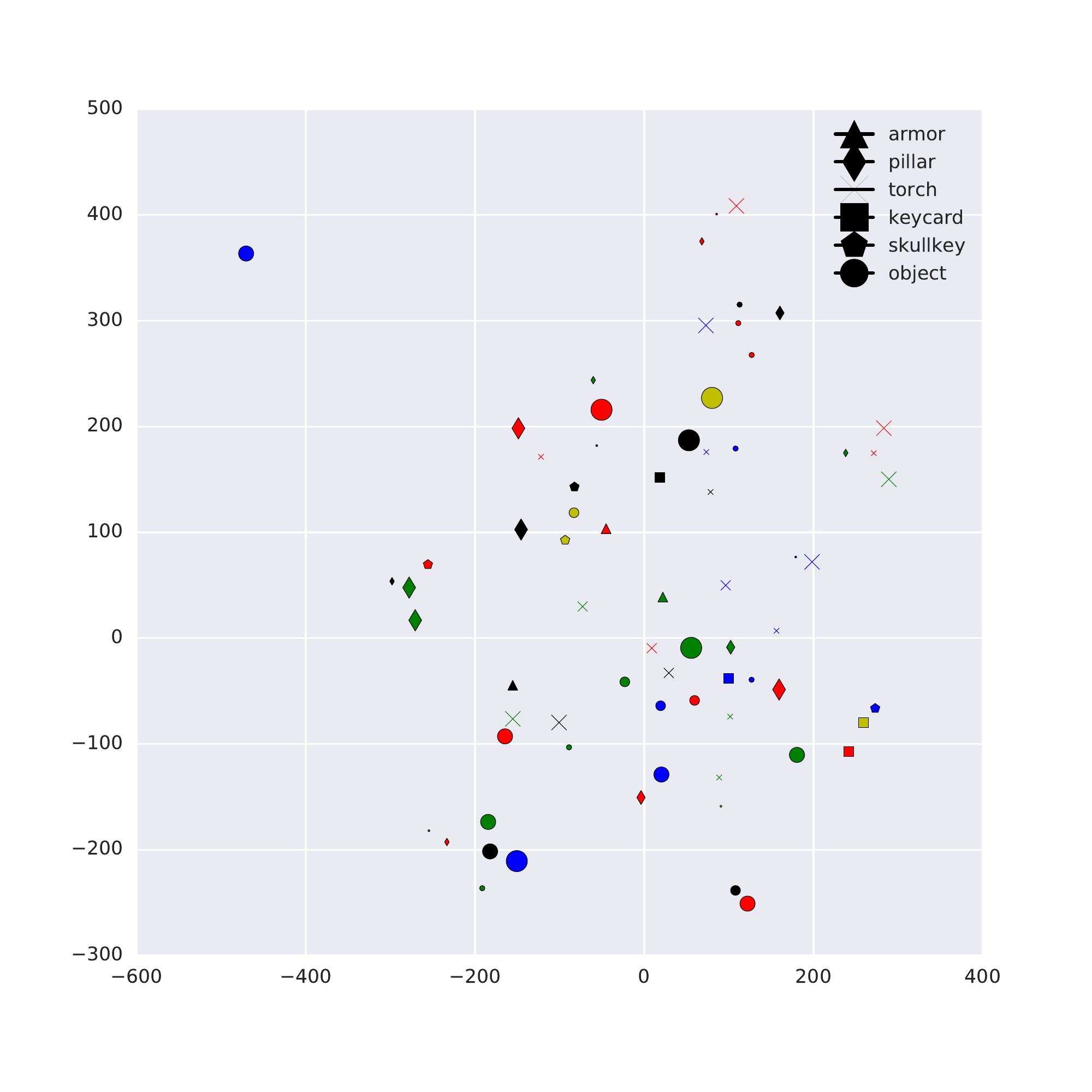}
		\end{subfigure}%
		\begin{subfigure}{.50\textwidth}
        \includegraphics[width=1.0\textwidth]{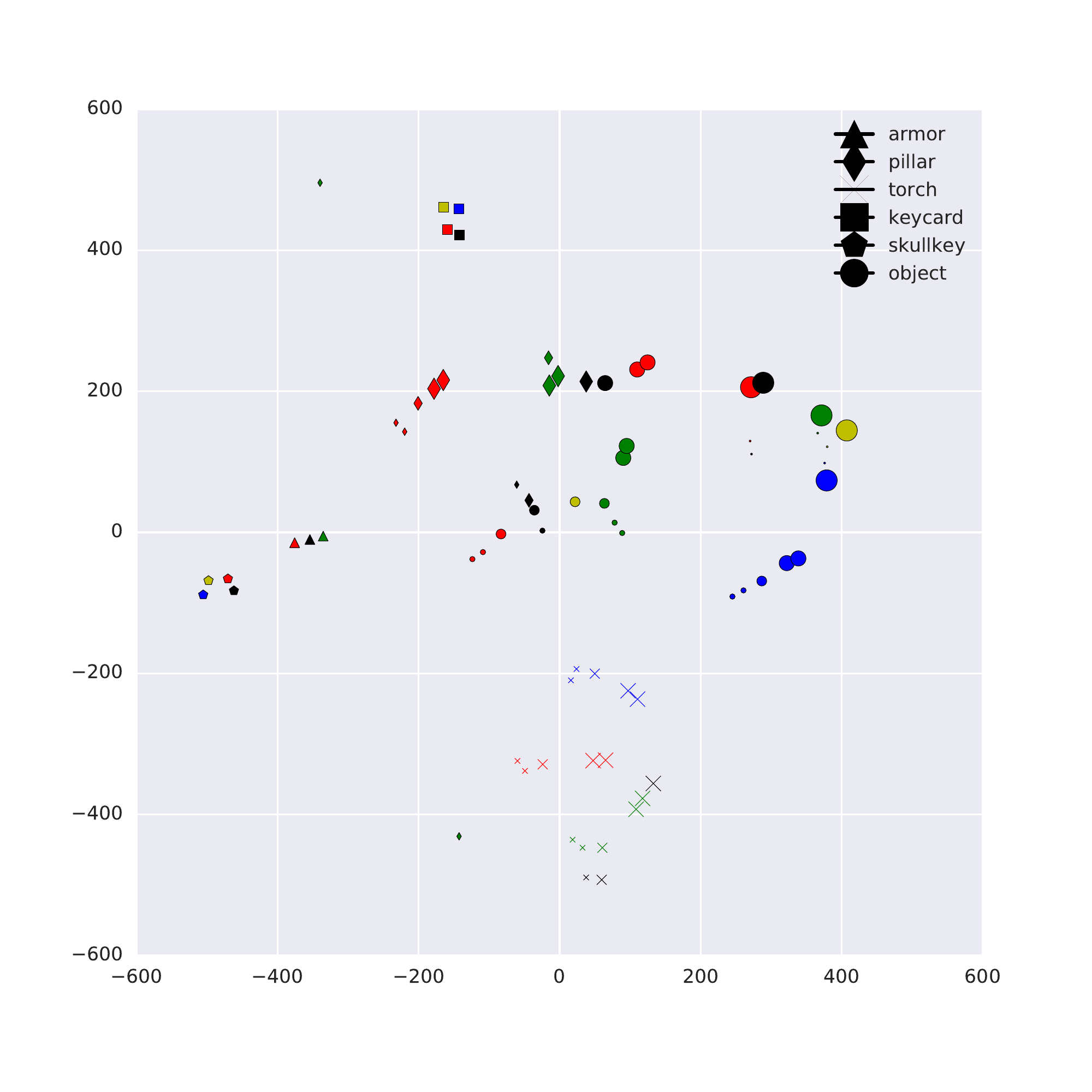}
        \includegraphics[width=1.0\textwidth]{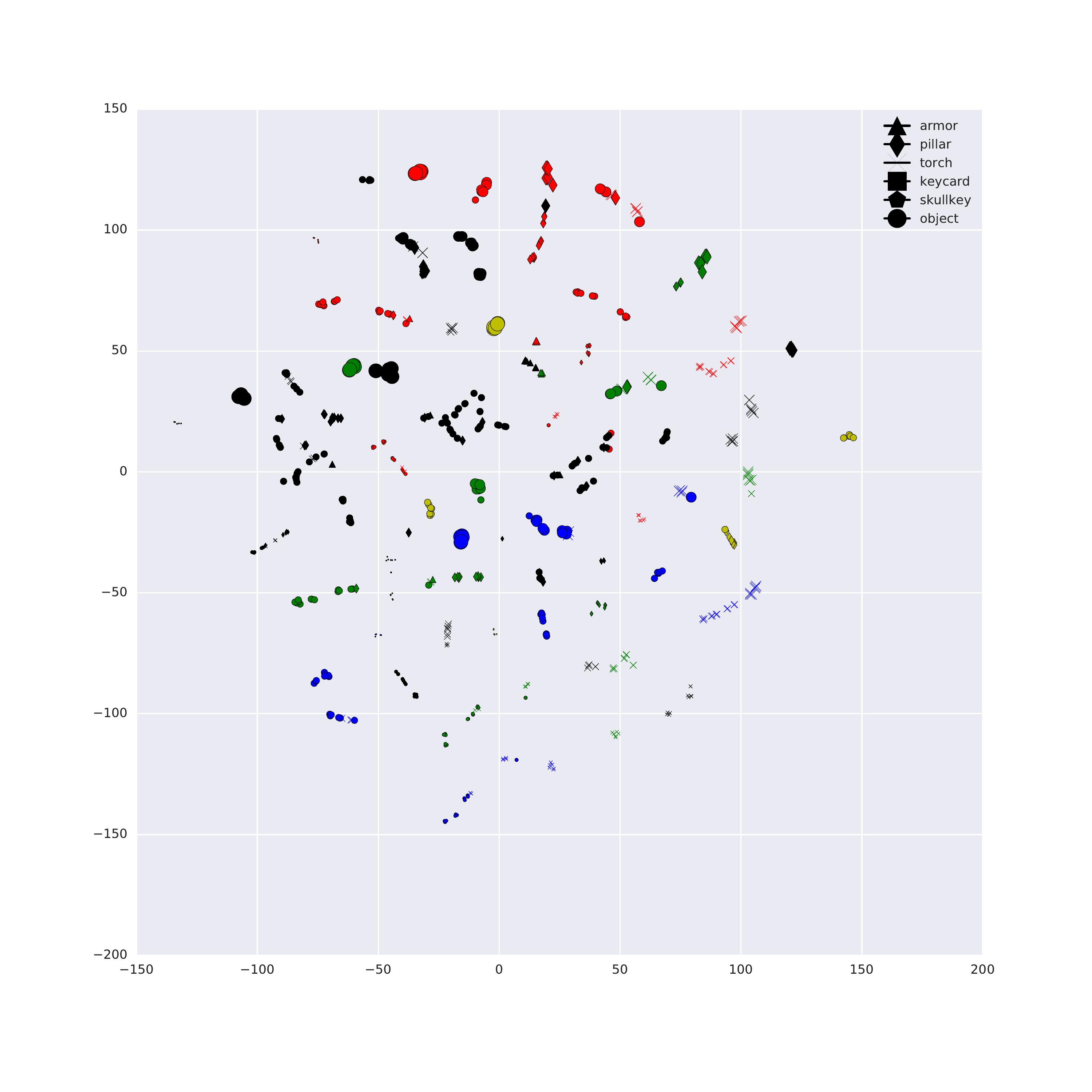}
		\end{subfigure}%
			
	\caption{t-SNE plots of instruction embeddings for GRU (top left), InferLite (top right), One-hot (bottom left) and Inferlite synonyms (bottom right). Best seen in electronic form.}
	\label{fig:tsne}
	\end{center}
\end{figure*}
\end{document}